%% file: main.tex
\documentclass{article}

\PassOptionsToPackage{numbers, compress}{natbib}

\usepackage[preprint]{neurips_2025}

\usepackage[utf8]{inputenc} 
\usepackage[T1]{fontenc}    
\usepackage{hyperref}       
\usepackage{url}            
\usepackage{booktabs}       
\usepackage{amsfonts}       
\usepackage{array}          
\usepackage{amsmath}        
\usepackage{nicefrac}       
\usepackage{microtype}      
\usepackage{xcolor}         
\usepackage{cleveref}
\usepackage[inline]{enumitem}
\usepackage{arydshln}
\usepackage{siunitx}  
\usepackage{graphicx} 
\usepackage[normalem]{ulem}     
\usepackage{etoolbox} 
\robustify\textbf     
\robustify\uline      
\usepackage{wrapfig}
\usepackage{caption}  
\usepackage{tabularx}
\usepackage{xspace}
\newcolumntype{L}[1]{>{\raggedright\arraybackslash\hspace{0pt}}p{#1}}
\newcolumntype{R}{>{\raggedright\arraybackslash}X} 

\newcommand{\eg}{\emph{e.g.}}
\newcommand{\ie}{\emph{i.e.}}
\newcommand{\vs}{\emph{vs.}}  

\title{Time-R1: Towards Comprehensive Temporal Reasoning in LLMs}

\author{%
  Zijia Liu, %
  Peixuan Han, %
  Haofei Yu, %
  Haoru Li, %
  Jiaxuan You %
  \vspace{2mm} 
  \\ 
  Siebel School of Computing and Data Science,
  University of Illinois at Urbana-Champaign\\
  \texttt{\{zliu331,jiaxuan\}@illinois.edu} 
}

\begin{document}

\maketitle

\vspace{-4mm}
\begin{abstract}
Large Language Models (LLMs) demonstrate impressive capabilities but lack robust temporal intelligence, struggling to integrate reasoning about the past with predictions and plausible generations of the future. Meanwhile, existing methods typically target isolated temporal skills, such as question answering about past events or basic forecasting, and exhibit poor generalization, particularly when dealing with events beyond their knowledge cutoff or requiring creative foresight. To address these limitations, we introduce \textit{Time-R1}, the first framework to endow a moderate-sized (3B-parameter) LLM with comprehensive temporal abilities: understanding, prediction, and creative generation. Our approach features a novel three-stage development path; the first two constitute a \textit{reinforcement learning (RL) curriculum} driven by a meticulously designed dynamic rule-based reward system. 
This framework progressively builds (1) foundational temporal understanding and logical event-time mappings from historical data, (2) future event prediction skills for events beyond its knowledge cutoff, 
and finally (3) enables remarkable generalization to creative future scenario generation without any fine-tuning. Strikingly, experiments demonstrate that Time-R1 outperforms models over 200 times larger, including the state-of-the-art 671B DeepSeek-R1, on highly challenging future event prediction and creative scenario generation benchmarks. This work provides strong evidence that thoughtfully engineered, progressive RL fine-tuning allows smaller, efficient models to achieve superior temporal performance, offering a practical and scalable path towards truly time-aware AI. To foster further research, we also release \textit{Time-Bench}, a large-scale multi-task temporal reasoning dataset derived from 10 years of news data, and our series of \textit{Time-R1} checkpoints.\footnote{Our code, the Time-Bench dataset, and Time-R1 model checkpoints are available at the project repository: \url{https://github.com/ulab-uiuc/Time-R1} and via our Hugging Face Collection: \url{https://huggingface.co/collections/ulab-ai/time-r1-682626aea47cb2b876285a16}.}
\end{abstract}

\input{1_introduction}
\input{2_related_work}
\input{3_method}
\input{4_experiments}
\input{5_discussion}

\input{6_conclusion}
\newpage
\section*{Appendix}
\addcontentsline{toc}{section}{Appendix} 

\appendix

\input{appendix}


\end{document}

%% file: 1_introduction.tex
\section{Introduction}


Large Language Models (LLMs) have achieved remarkable success across a spectrum of language understanding, generation, and even some complex reasoning tasks\cite{vaswani2017attention,brown2020language,kumar2025llm}. However, a persistent shortcoming in even the most advanced LLMs is their temporal reasoning ability\cite{chu2023timebench,yuan2024back}. This encompasses several key capacities\cite{bajpai2024temporally,zhou2020temporal,ding2024understanding}: accurately interpreting temporal relationships within their existing knowledge base (such as inferring event times, time differences, event order, and completing temporal entities), predicting the timing of future events based on learned patterns, and creatively generating plausible future events anchored in time. 
Studies have shown that most LLMs indeed struggle to update or contextualize knowledge under time constraints \cite{kim2025counterfactual}; even frontier models have been observed to perform worse than some smaller models in tasks that require integrating new temporal information \cite{wu2024updating}. This suggests a systemic weakness in how current LLMs grasp time. This weakness stems from multiple factors: architectural limitations \cite{nylund2023time}, such as the lack of explicit module representation of time; the static nature of their training corpora \cite{rae2021scaling}, which inevitably become outdated; and the non-chronological training process \cite{zhao2024set}, where temporal information across different periods is processed concurrently rather than sequentially, hindering the development of robust logical mappings between events and their corresponding times.

\begin{wrapfigure}{r}{0.5\textwidth}
\vspace{-1em}
    \centering
    \includegraphics[width=\linewidth]{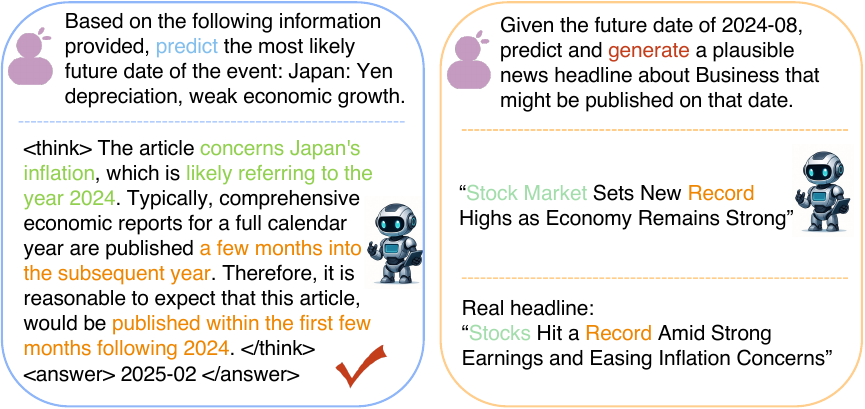} 
    \caption{\textbf{Generated outputs from Time-R1 showcasing its capabilities.} (Left) Future Event Time Prediction (Stage 2). (Right) Creative Scenario Generation (Stage 3), with output compared to a real-world headline.} 
    \label{fig:time_r1_examples}
    \vspace{-5mm} 
\end{wrapfigure}

While existing research aims to enhance temporal reasoning—for instance, Zhao \textit{et al.} \cite{zhao2024set} aligned LLM knowledge to target times, Kim \textit{et al.} \cite{kim2025counterfactual} improved temporal consistency, and Yuan \textit{et al.} \cite{yuan2024back} focused on future event prediction, with other works exploring representation methods \cite{su2024timo,xiong2024large}—these efforts often target isolated skills. They typically fall short of endowing LLMs with unified, comprehensive temporal intelligence that spans past understanding, future prediction, and creative, time-anchored generation, especially for events beyond their knowledge cutoffs \cite{zhao2024set,yuan2024back}.


In this paper, we aim to bridge this gap by equipping a single 3B-parameter model with comprehensive temporal reasoning capabilities through multi-stage Reinforcement Learning (RL), which has become a powerful framework for improving LLM reasoning. Recent frontior models such as OpenAI-o1 \cite{jaech2024openai} and DeepSeek-R1 \cite{guo2025deepseek} utilize RL methods like PPO \cite{schulman2017proximal} and GRPO \cite{shao2024deepseekmath}, proving effectiveness to learn complex reasoning capabilities, such as mathematical problem solving and multi-step logical deduction. 
We build upon Qwen2.5-3B-Instruct, a moderate-sized LLM, and demonstrate that through specialized training it can surpass models over 200× larger (for instance, DeepSeek-R1, a 671B-parameter model) on highly challenging temporal prediction and generation tasks. We propose a three-stage framework with RL and dynamic rewards to progressively establish the model's unified temporal capabilities, spanning temporal logic, future prediction, and time-anchored scenario generation: 
\textbf{(1) Stage 1 - Comprehension}: RL fine-tune the model using pre-cutoff data from a cold start on four fundamental temporal tasks – timestamp inference, time-difference estimation, events ordering, and masked time entity completion – to develop powerful logical mappings between events and their corresponding times.
\textbf{(2) Stage 2 - Prediction}: Further train the model to predict events occurring after knowledge cutoff, thereby teaching it to utilize general reasoning ability built in Stage 1 to extrapolate trends and anticipate future outcomes.
\textbf{(3) Stage 3 - Generation}: Directly have the model generate logical future scenario without fine-tuning, leveraging the capabilities obtained from the first two stages.

Through this staged curriculum, the LLM thus progresses from comprehending known temporal facts to skillfully navigating the complexities of the future. This advanced training culminates in robust capabilities for both predicting future event timelines and creatively generating plausible scenarios for unseen future contexts---addressing significant limitations in how current AI handles such challenging forward-looking tasks. Illustrative examples of these advanced future-oriented skills, such as Time-R1's proficiency in forecasting event dates and generating contextually appropriate news headlines for future dates (as depicted in \Cref{fig:time_r1_examples}), highlight the practical efficacy of our approach. 

In summary, the key contributions of our work are as follows: \textbf{(1) Unified Temporal Reasoning in One Model:} We introduce the first LLM that exhibits a holistic temporal reasoning ability encompassing logic, 
prediction, and generation. \textbf{(2) Small Model, Big Performance:} We show that a relatively small 3B model, when fine-tuned with our meticulously designed multi-stage dynamic-reward RL strategy, can match or even exceed the performance of models with hundreds of billions of parameters (\eg, the 671B-parameter R1 model) on temporal prediction and generation tasks. \textbf{(3) Fast Adaptability and Cost Efficiency:} Our approach demonstrates that temporal knowledge can be continuously refreshed in a cost-effective manner. A 3B model can be quickly fine-tuned on new data as time progresses, which is infeasible for a hundreds of billion model that would require enormous computational resources (on the order of millions of dollars for fine-tuning). 
\textbf{(4) Resources for the Community:} To encourage further research in temporal-aware AI, we release \textbf{Time-Bench}, a dataset of over 200,000 examples with explicit temporal annotations covering diverse tasks including timestamp inference, time-gap estimation, event ordering, and temporal entity completion. We also release \textbf{Time-R1}, a series of high-performing and continuously updatable temporal reasoning model checkpoints, offering a strong foundation for future time-aware LLM development and iterative refinement.

%% file: 2_related_work.tex
\section{Related Work}

\textbf{Temporal Reasoning in LLMs.} While adept at many complex tasks \cite{guo2025deepseek,guo2024deepseek}, LLMs struggle significantly with temporal reasoning—understanding time and event interrelations—a faculty crucial for comprehensive world understanding and interaction \cite{chu2023timebench,deroy2024short,bajpai2024temporally}.
Recent studies increasingly target these deficiencies, often focusing on specific temporal facets. For example, some efforts aim to improve temporal accuracy by aligning LLM knowledge with a target time for time-sensitive questions \cite{zhao2024set}. Meantime, some investigate methods for better integrating temporal information into model representations \cite{su2024timo}, while others explore leveraging external knowledge sources or structured representations like temporal graphs to augment LLM capabilities \cite{xiong2024large}. However, LLMs exhibit particularly poor generalization when reasoning about the future, especially for events beyond their knowledge cutoff or tasks requiring creative foresight. Consequently, robust methods for direct, challenging future event prediction or creative scenario generation remain scarce in the literature.
While some initiatives explore future event prediction and forecasting (e.g., Yuan \textit{et al.} \cite{yuan2024back} employed instruction tuning to predict event occurrences from past contexts), comprehensive approaches addressing the full spectrum of complex and creative future-oriented reasoning are largely underdeveloped.

\textbf{Reinforcement Learning in LLMs.} Reinforcement learning (RL) has recently attracted attention due to its scalability and enhanced generalization capabilities. Building on policy optimization algorithms like PPO \cite{schulman2017proximal}, reinforcement learning from human feedback (RLHF) — the first application of RL to large language models — has become a standard paradigm for aligning LLMs with desired behaviors \cite{ouyang2022training,kaufmann2023survey}. Recent advances aim to simplify or improve this process: Direct Preference Optimization (DPO) \cite{rafailov2023direct} and Simple Preference Optimization (SimPO) \cite{meng2024simpo} replace the conventional RL loop with more direct optimization of preference-based rewards, eliminating the need for a separate reward model or reference policy. Other methods are tailored specifically for LLMs; for instance, Group Regularized Policy Optimization (GRPO) \cite{shao2024deepseekmath} introduces a group-based reward formulation in place of a single critic, achieving more stable training and better generalization. Likewise, Ahmadian \textit{et al.} \cite{ahmadian2024back} revisit classic policy gradient techniques \cite{williams1992simple} to propose RLOO (REINFORCE-Leave-One-Out), an online RL algorithm that refines LLM policies with reduced variance and cost. These RL-driven approaches have demonstrated notable gains in LLM reasoning capabilities. In particular, GRPO and related strategies have yielded state-of-the-art performance on complex reasoning tasks including mathematical problem solving \cite{shao2024deepseekmath,xie2025logic}, search engine interaction and knowledge retrieval \cite{jin2025search,song2025r1}, code generation tasks \cite{li2025torl} and others \cite{qian2025toolrl,wang2025otc,chen2025rm}. Despite these successes, the application of reinforcement learning to temporally-grounded reasoning remains underexplored. 
This gap suggests an opportunity to leverage RL methods to develop unified, time-sensitive reasoning abilities in future LLMs.

%% file: 3_method.tex
\section{Method}\label{sec:method}


This section details the \textbf{Time-R1} methodology for enhancing LLM temporal capabilities via Reinforcement Learning (RL) fine-tuning. We introduce a novel three-stage training framework (\Cref{sec:experimental_design}) guided by a dynamic, rule-based reward system (\Cref{sec:rewards}). We first outline the underlying RL optimization setup using Group Relative Policy Optimization (GRPO) (\Cref{sec:problem}) before detailing these core framework and reward components.

\subsection{Reinforcement Learning Fine-tuning for Temporal Reasoning}
\label{sec:problem}


Our approach employs reinforcement learning (RL) to fine-tune a Large Language Model (LLM) for complex temporal reasoning tasks. The core process involves interaction between the LLM policy and a rule-based environment. Given a prompt $x$ detailing a specific temporal task, the LLM, parameterized by $\theta$, generates an output sequence $y$ autoregressively according to its current policy $\pi_\theta(y\mid x)=\prod_{t=1}^{|y|}\pi_\theta(y_t\mid x,y_{<t})$.

\begin{figure}[t!]
    \centering
    \includegraphics[width=0.98\textwidth, keepaspectratio]{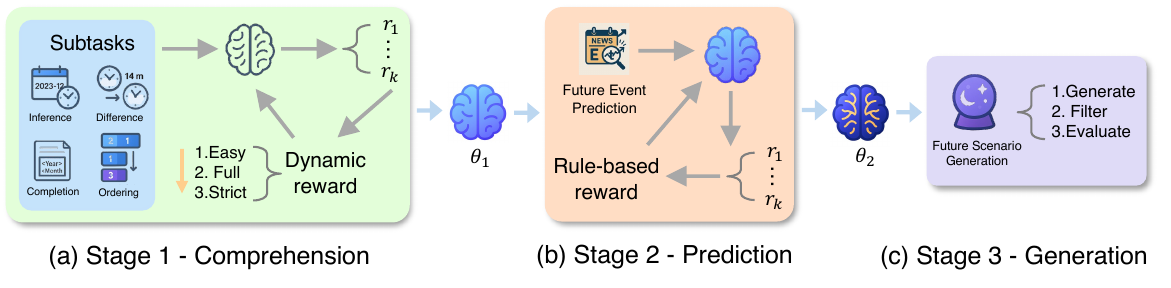} 
    \caption{\textbf{Overview of the Time-R1 framework.} 
    The process consists of three stages: (a) Stage 1 establishes foundational understanding by fine-tuning a base LLM on historical data across four temporal subtasks, driven by reinforcement learning (GRPO) and a dynamic reward system, resulting in model $\theta_1$. (b) Stage 2 trains $\theta_1$ for future event time prediction using post-cutoff data and a rule-based reward, producing $\theta_2$. (c) Stage 3 leverages $\theta_2$ for inference-based creative future scenario generation, followed by evaluation, without further RL.}
    \label{fig:method}
\end{figure}

\textbf{Structured Generation Process.} 
To facilitate complex reasoning, interpretability and structured output, we guide the model generation process. For all tasks, the LLM is prompted using specific templates incorporating system instructions (\ie, instructing the model to reason first: ``You are a helpful assistant. You first think about the reasoning process in your mind and then provide the user with the answer.'') to generate its reasoning within ``<think>...</think>'' tags, followed by the final answer within ``<answer>...</answer>'' tags. The entire generated sequence $y$, encompassing both thought and answer components, constitutes the output evaluated by the environment.

\textbf{Policy Optimization using GRPO.}
\label{sec:grpo} 
The environment evaluates the output $y$ using a task-specific dynamic reward function $R(x, y)$ (detailed in \Cref{sec:rewards}). To optimize the policy parameters $\theta$, we utilize Group Relative Policy Optimization (GRPO) \cite{shao2024deepseekmath}. A key challenge in RL fine-tuning of LLMs is the high variance often associated with policy gradient estimates\cite{sutton1998reinforcement}. GRPO addresses this by calculating the advantage of a generated response relative to other responses sampled for the same input prompt, thereby providing a more stable learning signal without requiring an auxiliary value function.

Specifically, for a given prompt $x$, we first sample a batch of $K$ responses $\{y_k\}_{k=1}^K$ using a reference policy $\pi_{\text{ref}}$ (typically the policy before the update step). After computing the reward $R(x, y_k)$ for each response, the group-normalized advantage $\hat{A}(x, y_k)$ for response $y_k$ is calculated as:
\begin{equation}
    \hat{A}(x, y_k) = R(x, y_k) - b(x), \,\,\,\, \text{where} \,\, b(x) = \frac{1}{K} \sum_{j=1}^K R(x, y_j).
    \label{eq:grpo_advantage}
\end{equation}
This advantage estimate $\hat{A}(x, y_k)$ reflects the relative quality of response $y_k$ compared to the average performance within its group.

To update the policy $\pi_\theta$ stably using this advantage, we employ a clipped surrogate objective function, similar in structure to that used in PPO \cite{schulman2017proximal}, which helps prevent large, detrimental policy updates. Let the probability ratio be $r_k(\theta) = \frac{\pi_\theta(y_k|x)}{\pi_{\text{ref}}(y_k|x)}$. The per-sample clipped objective term is:
\begin{equation}
    L_k^{\text{CLIP}}(\theta) = \min \left( r_k(\theta) \hat{A}(x, y_k), \,\, \text{clip}\left(r_k(\theta), 1-\epsilon, 1+\epsilon \right) \hat{A}(x, y_k) \right)
    \label{eq:ppo_clip}
\end{equation}
where $\epsilon$ is the clipping hyperparameter. The overall objective function $J_{\text{GRPO}}(\theta)$ maximized during training balances the expected clipped advantage with a KL-divergence penalty against the reference policy $\pi_{\text{ref}}$:
\begin{equation}
  \max_{\theta} J_{\text{GRPO}}(\theta) = 
    \mathbb{E}_{x \sim \mathcal{D}, \{y_k\} \sim \pi_{\text{ref}}} [ \frac{1}{K} \sum_{k=1}^K L_k^{\text{CLIP}}(\theta) ]
    \;-\;
    \beta\,\mathbb{E}_{x\sim\mathcal{D}} \mathbb{D}_{\mathrm{KL}}
    [\pi_\theta(\cdot\mid x)\;\|\; \pi_{\text{ref}}(\cdot\mid x)],
  \label{eq:grpo_objective_final} 
\end{equation}
where $\mathcal{D}$ is the training dataset union, $\beta$ controls the KL penalty strength, $\mathbb{D}_{\mathrm{KL}}$ is the Kullback–Leibler divergence, and $\pi_{\text{ref}}$ is the stage-specific frozen reference policy (initialized from Qwen2.5-3B-Instruct for Stage 1) used for both advantage calculation reference and KL regularization. This objective guides the policy towards higher rewards, leveraging the stable GRPO advantage estimates within a constrained optimization framework.


\subsection{Time-R1: A Three-Stage Temporal Learning Framework}
\label{sec:experimental_design}

To empirically evaluate the effectiveness of our proposed methodology (outlined in \Cref{sec:problem}), we designed a comprehensive three-stage experimental procedure to train \textbf{Time-R1}, as shown in \Cref{fig:method}. This staged approach aims to progressively cultivate sophisticated temporal logic, prediction, and generation capabilities within the Large Language Model (LLM). We detail each stage below. 

\subsubsection{Stage 1 - Comprehension: Foundational Temporal Understanding via RL Fine-tuning}
\label{sec:stage1}

\textbf{Objective.} The primary goal of this initial stage is to establish a robust foundation for temporal comprehension within the LLM. We aim to instill the ability to interpret fundamental temporal relationships between events and their corresponding times by fine-tuning the model using historical news data from \textit{before} its knowledge cutoff date. 

\textbf{Dataset.} We construct a specialized dataset derived from a large corpus of New York Times (NYT) news articles \cite{NYTArchiveAPI} (over 200,000) spanning eight years, from January 2016 to December 2023. We extract the headline $h$ and abstract $a$ of the news article to represent each event $E$, \ie, $E=(h,a)$. Details can be found in Appendix \ref{app:nyt_corpus_details}. 

\textbf{Subtasks.} From this corpus, we curate data instances tailored to four specific and fundamental temporally-focused and logic-based subtasks \cite{pustejovsky2003timeml,gast2016enriching}:
\textbf{(1) Timestamp Inference:} Infer the specific date $t$ (\eg, 2023-12) associated with a described event $E$. \textbf{(2) Time-Difference Estimation:} Estimate the temporal gap $\Delta t$ (\eg, 14 months) between two described events, $E_1$ and $E_2$.
\textbf{(3) Event Ordering:} Determine the correct chronological sequence $C$ (\eg, Event order: 2-1-3) of three events $E_1$, $E_2$ and $E_3$ presented out of order.
\textbf{(4) Masked Time Entity Completion:} Fill in a masked temporal expression $M_{e}$ (\ie, <Year> and <Month>) within a given event description $E'$.

In order to help the model develop general logic and indeed acquire the skill to accurately map events to their respective times from textual clues, we force the model to infer each event's date first and then give a task-specific answer for every subtask except the first. Both would be judged a score that would then serve as a part of the reward (see \Cref{sec:rewards}).  
Consequently, this prevents the model from merely guessing the final answer implicitly. For instance, for the Masked Time Entity Completion task, success hinges on the model's ability to discern detailed semantics from the surrounding text. This is crucial because the specific temporal entity to be completed often refers to a time distinct from the primary date of the event itself, thus pushing the model beyond simple date extraction towards a deeper contextual understanding to answer both correctly. By mastering these diverse subtasks, the LLM (\ie, a model checkpoint, denoted $\theta_1$) builds a robust foundational temporal understanding.

\subsubsection{Stage 2 - Prediction: Future Event Time Prediction via RL Fine-tuning}
\label{sec:stage2}

\textbf{Objective.} After obtaining the foundational capabilities developed in Stage 1, the objective of Stage 2 is to further train the model to predict the timing of future events occurring \textit{after} its initial knowledge cutoff (2023). This involves teaching the model to recall relevant and similar events in the past and their occurrence dates, extrapolate learned temporal development patterns and anticipate future event occurrences based on emerging, post-cutoff information.

\textbf{Dataset.}
For Stage 2, the training dataset, denoted $\mathcal{D}_{\text{train}}^{\smash{(2)}}$, is meticulously constructed to facilitate fair evaluation and strictly prevent data leakage from the test period.
To ensure a level playing field and align with the knowledge cutoff of the latest baseline models (e.g., DeepSeek-V3-0324-671B with a knowledge cutoff in July 2024), we first incorporate real news data. Specifically, we include a corpus of 7,000 real news articles from January 2024 to July 2024.
To train for predicting events beyond this cutoff (August 2024 - February 2025) without using real data from this period, we employ a data synthesis strategy. 
The synthetic dataset, created using the DeepSeek-V3 model informed by news from May to July 2024, constitutes approximately only half the volume of the real news data used for the earlier months.
This approach of using exclusively synthetic data for the future period 
is a deliberate measure to strictly avoid any potential data leakage, as the test dataset $\mathcal{D}_{\text{test}}^{\smash{(2)}}$ is real news events from this period (August 2024 - February 2025). Further details about the datasets can be found in Appendix \ref{app:synthetic_data_generation}. 

\textbf{Task.} In this stage, the model predicts the specific date $t$ for a news event $E$ based on its extracted headline $h$ and abstract $a$.

Initializing the model with the checkpoint $\theta_1$ obtained from Stage 1, we continue the fine-tuning process using GRPO on post-cutoff news while carefully controlling the information availability to simulate a true “future prediction” scenario. After training, this stage addresses the challenge that LLMs normally cannot generalize to events post-training \cite{lee2023temporal} and results in another model checkpoint, $\theta_2$, specialized in future event time prediction.

\subsubsection{Stage 3 - Generation: Creative Future Scenario Generation and Evaluation}
\label{sec:stage3}

\textbf{Objective.} In the final stage, we pivot from training to application – aiming to leverage the logical and predictive capabilities instilled in Stages 1 and 2 to enable the fine-tuned model to directly generate plausible, diverse, and temporally coherent future scenarios. This moves beyond predicting specific event times to creatively generating descriptions of hypothetical events given a specific future date. 

\textbf{Methodology.} This stage utilizes the model checkpoint $\theta_2$, obtained from Stage 2, exclusively for inference without any further RL fine-tuning. The process involves three sequential steps: future news generation, diversity-based filtering, and plausibility evaluation against real news.

First, the model generates hypothesized news events for specified future months $M$ (\ie, July 2024 onwards). To ensure comprehensive topical coverage, generation is conditioned on $T=8$ common and distinct themes $\tau$ (\eg, Foreign Affairs, Business, Technology, Politics). 
To enhance the richness of the output pool, each prompt asks the model to create multiple unique news (\ie, 3). This process results in a raw set of generated news items $\mathcal{G}_{\text{raw}}$ including each month $m$ and theme $\tau$.

Second, to curate a varied and non-redundant set of scenarios for evaluation, a diversity filtering process is applied to the raw generated articles $\mathcal{G}_{\text{raw}}$. We compute semantic embeddings $\mathbf{g} \in \mathbb{R}^{384}$ for each generated item $g$ using all-MiniLM-L6-v2 encoder \cite{wang2020minilm}, which retains excellent semantic capture capabilities through knowledge distillation from larger models \cite{galli2024performance}. Within each theme $\tau$ and month $m$, a greedy selection algorithm iteratively constructs a diverse subset. 
This filtering yields a curated set $\mathcal{G}_{\text{filt}, m}$ containing $N_{\text{div}}=5$ high-diversity news items per theme per month, totaling $N_g = T \times N_{\text{div}} = 40$ representative generated scenarios for each month $m$. 

Finally, the realism and plausibility of the generated future scenarios are quantified through comparison with actual news events from the corresponding future months. The ground truth consists of real news events $r$ from the held-out test dataset $\mathcal{D}_{\text{test}}^{\smash{(2)}}$, partitioned by month $m$ into sets $\mathcal{D}_{\text{real}, m}$. 
We compute semantic embeddings $\mathbf{A}_g$ for the filtered generated news items $g \in \mathcal{G}_{\text{filt}, m}$ and $\mathbf{B}_r$ for the real news items $r \in \mathcal{D}_{\text{real}, m}$, using the same ``all-MiniLM-L6-v2'' model. The semantic relatedness between a generated item $\mathbf{A}_g$ and a real item $\mathbf{B}_r$ is measured using cosine similarity: $\text{sim}(\mathbf{A}_g, \mathbf{B}_r) = \cos(\phi) = \frac{\mathbf{A}_g \cdot \mathbf{B}_r}{\|\mathbf{A}_g\| \|\mathbf{B}_r\|}$, 
where $\phi$ represents the angle between the 384-dimensional embedding vectors. To assess overall plausibility for a given month $m$, we calculate the Average Maximum Similarity (AvgMaxSim) score. For each generated news item $\mathbf{A}_{g,i}$ ($i=1, \dots, N_g$), we find its maximum similarity to any real news item in that month, $\max_{\mathbf{B}_r \in \mathcal{D}_{\text{real}, m}} \text{sim}(\mathbf{A}_{g,i}, \mathbf{B}_r)$. The AvgMaxSim score is the average of these maximum similarity values across all $N_g$ generated items:
\begin{equation}
\label{eq:avgmaxsim}
\text{AvgMaxSim}_m = \frac{1}{N_g} \sum_{i=1}^{N_g} \left( \max_{\mathbf{B}_r \in \mathcal{D}_{\text{real}, m}} \text{sim}(\mathbf{A}_{g,i}, \mathbf{B}_r) \right)
\end{equation}
This metric quantifies, on average, how closely the generated plausible future events align semantically with events that actually transpired during that month. The process culminates in generating monthly AvgMaxSim reports and visualizations, facilitating quantitative comparisons against baseline generative models or ablations of our framework. 

In summary, Stage 3 serves as evidence for the generalization fostered by our first two stages  RL framework. It reveals that the strong temporal grounding comprehension and predictive skills learned previously, combined with the LLM's innate linguistic abilities, readily and effectively generalize, allowing the model to anticipate future event dynamics and generate plausible, creative scenarios accordingly, without task-specific fine-tuning for this generative capability.

\subsection{Reward Design}
\label{sec:rewards}

A meticulously engineered reward function, $R(x, y)$, underpins the success of our Time-R1 framework. Its comprehensive and rigorous design, refined through iterative experimentation, has proven critical for developing the nuanced temporal reasoning abilities observed in our model (see experimental validation in \Cref{sec:experiments}, detailed analysis in \Cref{sec:discuss}, and more illustration in Appendix). 
The reward function $R(x, y)$ serves as the primary training signal guiding the policy optimization process outlined in \Cref{eq:grpo_objective_final}. We adopt a rule-based dynamic reward system that assesses the correctness and quality of the model's generated output $y$ given the prompt $x$. The final scalar reward $R(x,y)\in[-0.8,\,1.1]$ incorporates several components: task-specific accuracy ($R_{\text{acc}}$), format rewards ($R_{\text{format}}$), and penalties ($P_{\text{penalty}}$) for undesirable outputs, \ie,
\begin{equation}
\label{eq:reward}
R(x,y) = R_{\text{acc}} + R_{\text{format}} - P_{\text{penalty}}
\end{equation}

\subsubsection{Universal Bonuses and Penalties Design}
\label{sec:universal_bonuses_penalties}

\textbf{Output Parsing and Format.}
We first parse the content $y_{\text{ans}}$ within the ``<answer>...</answer>'' tags. If $y_{\text{ans}}$ is missing or contains explicit refusal terms like ``no event'' or ``none'', a penalty $P_{\text{no\_event}}$ is applied (\ie, $P_{\text{no\_event}}\! \in \{0.1, 0.2\}$ for Stage 1 tasks, and $\{0.2, 0.3\}$ for Stage 2 prediction, depending on severity).

\textbf{Common Bonuses and Penalties.}
A set of bonuses and penalties apply across tasks to encourage well-formed and concise outputs:
\begin{itemize}
    \item \textbf{Format Adherence Bonus (}$R_{\text{ans\_fmt}}$\textbf{):} A small bonus $b_{fmt} = 0.05$ is awarded if the content $y_{\text{ans}}$ adheres to the expected format for the specific task (\eg, ``YYYY-MM'' format for date inference, and specific structures for multi-part answers). Valid format is also a prerequisite for accuracy scoring. Range: $\{0, 0.05\}$.
    \item \textbf{Tag Structure Bonus (}$R_{\text{tags}}$\textbf{):} Minor bonuses ($b_{tag} = 0.025$) are given for both the correct presence and count of structural tags (\eg, ``<think>'', ``</answer>''), incentivizing the chain-of-thought structure. Range: $[0, 0.05]$.
    \item \textbf{Length and Repetition Penalty (}$P_{\text{len\_rep}}$\textbf{):} A penalty is subtracted to discourage overly verbose or repetitive outputs; this mechanism has proven particularly effective in our empirical experiments (see cases in \Cref{Appendix:cases,app:repete}). Range: $[0,\,0.5]$.
    \begin{equation}
P_{\text{len\_rep}} = \max(\,P_{\text{length}}, P_{\text{repetition}})
\end{equation}
where $P_{\text{length}}$ penalizes responses (of $N$ tokens) exceeding a length threshold $L_{thresh}$ (\ie, 900 tokens) to prevent them from approaching the maximum allowed length $L_{max}$ (\ie, 1024 tokens). This is calculated as:
\begin{equation}
P_{\text{length}} = \min(1.0, \frac{N - L_{thresh}}{L_{max} - L_{thresh}}) \times 0.3, \quad \text{if } N > L_{thresh}
\end{equation}
$P_{\text{repetition}}$ is the maximum of three distinct repetition penalties:
\begin{equation}
P_{\text{repetition}} = \max(P_{\text{word\_repeat}}, P_{\text{phrase\_repeat}}, P_{\text{ngram\_diversity}})
\end{equation}
where $P_{\text{word\_repeat}}$ penalizes sequences of more than 5 identical consecutive words, $P_{\text{phrase\_repeat}}$ penalizes recurring phrases, and $P_{\text{ngram\_diversity}}$ penalizes insufficient global n-gram diversity. The combined penalty $P_{\text{repetition}}\! \in [0, 0.5]$.
\end{itemize}

\subsubsection{Task-Specific Accuracy Score.}
\label{sec:task_specific_rewards}
Accuracy score ($R_\text{acc}\!\in[0,\,1]$) is the core component of our reward mechanism, varying by task:

\textbf{Timestamp Inference:}
The task is to infer the date $t_p$ for a given event $E$. Let $t_{gt}$ be the ground truth date. The accuracy score is based on the temporal distance $\Delta m(t_p, t_{gt})$ (in months) between the inference and target:
\begin{equation}
    R_\text{acc} = R_{\text{date}}(t_p, t_{gt}, \alpha) = e^{(-\alpha \cdot \Delta m(t_p, t_{gt}))}
    \label{eq:reward_date}
\end{equation}
where $\alpha$ is a decay coefficient. For Stage 1 inference, $\alpha$ is dynamically adjusted based on sample difficulty and training step (ranging between $0.07$ and $0.1$). This exponential reward structure, particularly when coupled with the dynamic $\alpha$, ensures that the reward signal clearly reflects the proximity of the inferred date to the ground truth, effectively allowing the model to perceive the magnitude of its temporal error $\Delta m(t_p, t_{gt})$. See \Cref{sec:dynamic_alpha} and \Cref{sec:ablation_dynamic_reward} for more discussion. 

\textbf{Time-Difference Estimation:}
The task is to infer the dates $t_{p1}, t_{p2}$ of two events and their difference $\Delta t_p$ (in months). Let ground truths be $t_{gt1}, t_{gt2}, \Delta t_{gt}$. The reward combines accuracy on dates and the difference, weighted ($w_d=0.25, w_{\Delta t}=0.5$), and includes an inconsistency penalty:
\begin{equation}
    R_\text{acc} = (w_d R_{d1} + w_d R_{d2} + w_{\Delta t} R_{\Delta t}) \cdot P_\text{incon}
    \label{eq:reward_diff}
\end{equation}
where $R_{d1} = R_{\text{date}}(t_{p1}, t_{gt1}, \alpha_1)$ and $R_{d2} = R_{\text{date}}(t_{p2}, t_{gt2}, \alpha_2)$ are date accuracy, using dynamic $\alpha_1, \alpha_2$. $R_{\Delta t} = e^{(-\alpha_{\Delta t} \cdot |\Delta t_p - \Delta t_{gt}|)}$ denotes difference accuracy, where $\alpha_{\Delta t}=0.05$ if $\Delta t_p \ge 25$, otherwise $\alpha_{\Delta t} = 0.1$ or $(\alpha_1+\alpha_2)/2$ depending on the dynamic strategy process, to balance the reward and to encourage more robust estimation even when the model is dealing with events separated by large time differences.
The inconsistency penalty factor ($P_\text{incon} \in (0, 1]$) penalizes discrepancies between the explicitly inferred difference $\Delta t_p$ and the difference implied by the inferred dates $|t_{p2}-t_{p1}|$; this penalty is designed to ensure the internal logical consistency of the model's output. Let the error be $\Delta_\text{incon} = ||t_{p2}-t_{p1}| - \Delta t_p|$. Then $P_\text{incon} = e^{(-\alpha_\text{incon} \cdot \Delta_\text{incon})}$, where the decay $\alpha_\text{incon}$ is smaller for larger $\Delta t_p$ (base $\alpha_\text{incon}=0.1$, scaled down if $\Delta t_p \ge 25$). 
The learning dynamics of $P_\text{incon}$, illustrating the model's progressive adherence to this logical constraint, are presented in Appendix~\ref{app:stage1_curves}.

\textbf{Event Ordering:}

The task involves inferring dates $t_{p1}, t_{p2}, t_{p3}$ and the correct chronological order $C_p$ (permutation) for three events $E_1, E_2, E_3$. Let ground truths be $t_{gt1}, t_{gt2}, t_{gt3}, C_{gt}$. The reward combines accuracy on dates and the order, weighted ($w_d=0.2, w_\text{ord}=0.4$), and includes both an inconsistency penalty and a diversity penalty:
\begin{equation}
    R_\text{acc} = (w_d \sum_{i=1}^3 R_{di} + w_\text{ord} R_\text{order}) \cdot P_\text{incon} \cdot P_\text{div}
    \label{eq:reward_order}
\end{equation}
where $R_{di} = R_{\text{date}}(t_{pi}, t_{gti}, \alpha_i)$ for $i=1,2,3$ is date accuracy, using dynamic $\alpha_i$. $R_\text{order}$ represents order accuracy, calculated based on the number of correctly ordered pairs in $C_p$ compared to $C_{gt}$ (\ie, $R_\text{order}=N_\text{correct\_pair}/N_\text{total\_pair}$, where $N_\text{total\_pair}=3$). 
The inconsistency penalty factor ($P_\text{incon} \in \{0.2, 0.4, 0.7, 1.0\}$) penalizes if the inferred order $C_p$ contradicts the order implied by the inferred dates $t_{p1}, t_{p2}, t_{p3}$ (based on pairwise similarity), thereby ensuring the model's explicit ordering aligns with the chronology of its inferred event dates. 
The diversity penalty factor ($P_\text{div} \in \{0.2, 1.0\}$) penalizes trivial solutions where all inferred dates $t_{pi}$ are identical, or where dates are sequential (\eg, $t_{p3}-t_{p2}=t_{p2}-t_{p1}=1$) and the order is trivial (\eg, 1-2-3); this encourages the model to infer more varied and realistic event date distributions rather than collapsing to overly simplistic patterns. $P_\text{incon}$ and $P_\text{div}$ are both proven effective in empirical experiments (see Appendix~\ref{app:stage1_curves}).

\textbf{Masked Time Entity Completion:}
The task is to infer the date $t_p$ of an event $E'$ and a masked entity $M_{e\_p}$ (either Year or Month). Let ground truths be $t_{gt}, M_{e\_gt}$. The reward combines accuracy on the date and the entity, weighted ($w_d=w_e=0.5$):
\begin{equation}
    R_\text{acc} = w_d R_\text{date} + w_e R_\text{entity}
    \label{eq:reward_comp}
\end{equation}
where $R_\text{entity} = e^{(-3\alpha \cdot \Delta m_c)}$ denotes entity accuracy, using dynamic $\alpha$. When the masked entity is ``Month'', $\Delta m_c$ represents the circular difference of exact or variant month name to better capture the proximity, \ie, $\Delta m_c=\min(|M_{e\_p} - M_{e\_gt}|, \,12 - |M_{e\_p} - M_{e\_gt}|)$.


\textbf{Future Event Prediction:} 
Similar to the Timestamp Inference task but for future events, however, this task employs a stricter evaluation standard as the model already has foundational temporal comprehension. 
Thus, the decay coefficient is a fixed larger value (\ie, $\alpha=0.1$) in \Cref{eq:reward_date}, resulting in more severe penalties for prediction errors. 

\subsubsection{Dynamic Reward Mechanism} 
\label{sec:dynamic_alpha}

To address the cold-start challenge inherent in fine-tuning LLMs for specialized temporal tasks and to foster robust performance\cite{xie2025logic}, particularly on more difficult examples, we employ a dynamic reward mechanism specifically during the Stage 1 RL fine-tuning process ( more discussion can be found at \Cref{sec:ablation_dynamic_reward}). This mechanism utilizes curriculum learning principles by adaptively adjusting the decay coefficient $\alpha$ used in the date accuracy reward component (\Cref{eq:reward_date}) based on data difficulty and training progression. This dynamic adjustment applies whenever $R_{\text{date}}$ is calculated for any Stage 1 subtask involving date inference (\ie, all four subtasks).

First, we stratify the Stage 1 training dataset based on difficulty. Using an initial model checkpoint (\ie, Qwen2.5-3B-Instruct), 
we perform Timestamp Inference task for all training samples. Samples where the absolute error in months ($\Delta m$) is less than or equal to 3 ($\Delta m \le 3$) are classified as ``easy'' level, while the remainder are classified as ``normal/hard''. 

The curriculum then proceeds in three sequential training steps, each building upon the model checkpoint from the previous step:

\textbf{Phase 1: Foundational Logic and Format Learning.} 
Initially, fine-tuning focuses exclusively on the Timestamp Inference task using only the samples classified as easy. During this step, we employ a fixed, relatively strict decay coefficient $\alpha = \alpha_{\text{target}} = 0.1$ in \Cref{eq:reward_date}. The primary goal is to enable the model to rapidly learn the fundamental task logic, establish correct response formatting, and build a solid foundation before encountering more complex tasks or difficult samples.

\textbf{Phase 2: Exploration on Full Task Suite.} 
Next, training expands to encompass all four Stage 1 subtasks and utilizes the full dataset (easy, normal, hard samples). For samples classified as normal/hard, we apply a lower, fixed decay coefficient $\alpha = \alpha_{\text{start}} = 0.07$. This more lenient penalty function encourages the model to explore diverse reasoning pathways for challenging instances across all tasks without being excessively penalized for initial inaccuracies. Easy samples continue to be evaluated using the stricter $\alpha = 0.1$.

\textbf{Phase 3: Transition to Strict Evaluation.} 
Finally, while continuing to train on all tasks and difficulty levels, we progressively increase the evaluation strictness for the normal/hard samples. The decay coefficient $\alpha$ for these samples transitions linearly from $\alpha_{\text{start}}=0.07$ up to $\alpha_{\text{target}}=0.1$ over $s_{\text{transition}}=50$ 
steps within this training phase, after which it remains fixed at $\alpha_{\text{target}}=0.1$ for any subsequent steps. Let $s$ be the current training step within this phase. 
The adaptive alpha $\alpha_{\text{transition}}(s)$ for normal/hard samples, is calculated as:
\begin{equation} 
    \label{eq:dynamic_alpha} 
    \alpha_{\text{transition}}(s) = \alpha_{\text{start}} + (\alpha_{\text{target}} - \alpha_{\text{start}}) \cdot \min(1.0, s/s_{\text{transition}}) 
\end{equation}
This gradual tightening of the reward function encourages the model to refine its precision on more difficult examples, adapting it towards the stricter evaluation standard ($\alpha=0.1$). This step aims to cultivate high accuracy across the entire data distribution by the end of Stage 1.

Importantly, this dynamic $\alpha$ adjustment schedule is employed strictly during the Stage 1 training process. For all evaluations performed on the test datasets (across all stages where applicable), we consistently use a fixed decay coefficient $\alpha = 0.1$ for all samples to ensure stable and comparable assessment of model performance.

\subsubsection{Final Reward Calculation.}
In summary, the total score $R(x, y)$ for a given task is computed by summing the relevant accuracy score and bonuses, then subtracting penalties introduced above. Thus, \Cref{eq:reward} can be further expressed as:
\begin{equation}
R(x, y) = R_\text{acc} + R_{\text{ans\_fmt}} + R_\text{tags} - P_\text{no\_event} - P_\text{len\_rep}
\end{equation}
Aggregating the potential minimum and maximum values of these components yields a range of $[-0.8, 1.1]$ for the total score $R(x, y)$. 

%% file: 4_experiments.tex
\section{Experiments}
\label{sec:experiments}


\subsection{Datasets.}
We utilize the datasets constructed from the New York Times (NYT) as described in \Cref{sec:experimental_design}.

\subsection{Baselines}
\label{sec:baselines}

To rigorously evaluate the performance of \textbf{Time-R1}, we compare it against two categories of six baseline models:
(1) \textbf{Instruction-Tuned LLMs of Varying Scales:} Qwen2.5-3B-Instruct (the base model for Time-R1), Qwen2.5-7B-Instruct \cite{yang2024qwen2} and Llama-3.1-8B-Instruct \cite{grattafiori2024llama} (medium-scale models), and DeepSeek-V3-0324-671B \cite{liu2024deepseek} (an extra-large generalist foundation model). 
(1) \textbf{Specialized Reasoning LLMs:} DeepSeek-Distill-Qwen-32B (a larger model with a strong emphasis on reasoning), and DeepSeek-R1-671B \cite{guo2025deepseek} (recognized for its state-of-the-art performance on a wide array of complex reasoning benchmarks). This comparison helps determine whether advanced, broad reasoning skills on well-trained models even with exceptionally large-scale can inherently address complex temporal tasks.



\subsection{Experimental Setup}
\label{sec:exp_setup}

\textbf{Implementation.}
All our experiments build upon Qwen2.5-3B-Instruct \cite{yang2024qwen2}, a moderate size for fast adaptability and cost efficiency. We implement our three-stage RL fine-tuning framework using veRL framework \cite{sheng2024hybridflow}, adopting the GRPO algorithm detailed in \Cref{eq:grpo_objective_final}. All RL fine-tuning experiments were conducted on four NVIDIA A6000 GPUs.


\textbf{Hyperparameters.}
Key hyperparameters for the GRPO optimization include KL coefficient $\beta = 0.001$, and $K = 5$ rollout responses per prompt for group-normalized advantage estimation. The full configuration details can be found at Appendix \ref{app:config_details}. 

\subsection{Main Results}
\label{sec:main_results}

We now present the core experimental results, evaluating the performance of Time-R1 across its training stages against the established baselines. We specifically report on the performance of the model checkpoint after Stage 1 ($\theta_1$) for foundational tasks and the checkpoint after Stage 2 ($\theta_2$) for future prediction and scenario generation.

\subsubsection{Stage 1: Foundational Temporal Reasoning Performance}
\label{sec:Stage1_results}

\begin{figure}[b!]
\centering
\includegraphics[width=0.96\linewidth]{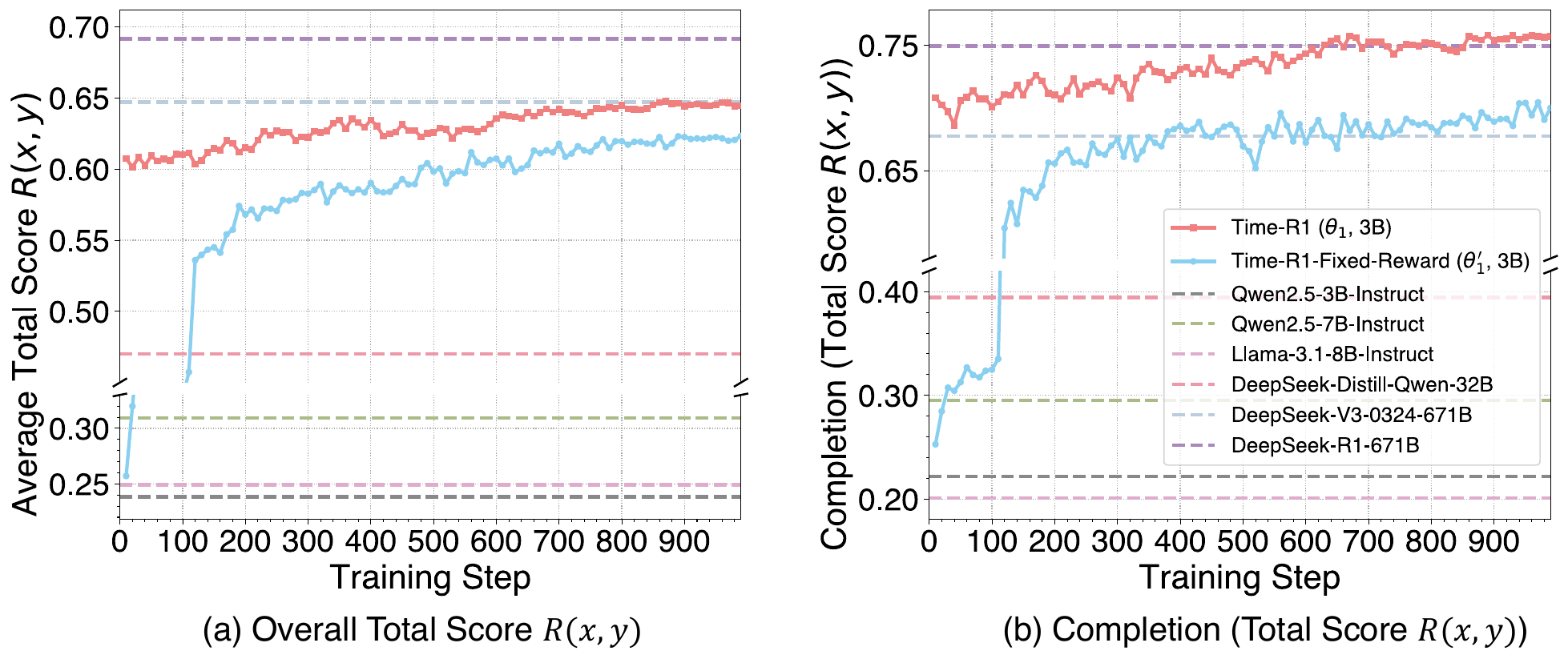}
\caption{\textbf{Stage 1 Training Performance \vs $\,$Baselines.} Training curves for Time-R1 ($\theta_1$) and its ablation variant, Time-R1-Fixed-Reward ($\theta_1'$), evaluated against baseline models (indicated by horizontal dashed lines). Plot (a) shows the Overall Total Score across all subtasks, while plot (b) presents the Masked Time Entity Completion subtask. The solid lines demonstrate our models' scores improving throughout the training process, ultimately surpassing the performance levels of most baseline models, including those with significantly larger scales.}
\label{fig:stage1_training_curves}
\end{figure}

The effectiveness of our Stage 1 fine-tuning on core temporal understanding is demonstrated by the training dynamics in \Cref{fig:stage1_training_curves} (see appendix~\ref{app:stage1_curves} for details of fine-tuning curves for all subtasks and phases) and the final scores in \Cref{tab:stage1_temporal_reasoning_avg_total_score_decimal}. The results highlight the substantial benefits of our Stage 1 RL fine-tuning. Time-R1 ($\theta_1$) demonstrates a remarkable improvement in its overall average score, with an increase of approximately 171.6\% over its base Qwen2.5-3B-Instruct model.

Significantly, with these improvements, Time-R1 now outperforms the much larger DeepSeek-V3-0324-671B model and is highly competitive with the state-of-the-art 671B DeepSeek-R1 model. It secures the top performance on the demanding Completion task and the second-best performance on the challenging Event Ordering task. This strong performance, rivaling or exceeding much larger baselines, is largely attributed to our meticulously designed task-specific rewards and the dynamic reward curriculum. For instance, the inconsistency and diversity penalties for Event Ordering (detailed in \Cref{sec:task_specific_rewards}) are pivotal. The learning curves in Appendix~\ref{app:stage1_curves} also illustrate that the model's adherence to response consistency and diversity for this task steadily improves, reflecting enhanced logical reasoning. Such effective instillation of logical mapping allows Time-R1 to compete effectively with much larger models on these complex temporal logic challenges. 

To validate the contribution of our reward design, we include an ablation model, \textbf{Time-R1-Fixed-Reward ($\theta_1'$)}, which was trained using a static, strict reward function. As shown in \Cref{fig:stage1_training_curves}, the full Time-R1 model consistently outperforms this ablation variant, underscoring the importance of the dynamic curriculum, which will be analyzed further in \Cref{sec:ablation_dynamic_reward}.

\begin{table}[t!]
\centering
\caption{\textbf{Stage 1 Foundational Temporal Reasoning Performance.} Average Total Score ($R(x,y)$) on the four subtasks and overall. Higher scores indicate better performance. Best score in each column is \textbf{bold}, second best is \uline{underlined}.}
\label{tab:stage1_temporal_reasoning_avg_total_score_decimal}
\resizebox{\textwidth}{!}{%
\begin{tabular}{l ccccc}
\toprule
\textbf{Model} & \textbf{Overall Avg.} $\uparrow$ & \textbf{Ordering} $\uparrow$ & \textbf{Completion} $\uparrow$ & \textbf{Inference} $\uparrow$ & \textbf{Difference} $\uparrow$ \\
\midrule
Qwen2.5-3B-Instruct & 0.2384 & 0.1583 & 0.2217 & 0.3372 & 0.2363 \\
Qwen2.5-7B-Instruct & 0.3092 & 0.2775 & 0.2953 & 0.3366 & 0.3275 \\
Llama-3.1-8B-Instruct & 0.2492 & 0.2239 & 0.2008 & 0.3339 & 0.2383 \\
DeepSeek-Distill-Qwen-32B & 0.4702 & 0.5026 & 0.3943 & 0.5264 & 0.4576 \\
DeepSeek-V3-0324-671B & 0.6471 & 0.6409 & 0.6777 & \uline{0.6796} & \uline{0.5901} \\
DeepSeek-R1-671B & \textbf{0.6916} & \textbf{0.6848} & \uline{0.7493} & \textbf{0.7145} & \textbf{0.6172} \\
\midrule
\textbf{Time-R1-Fixed-Reward ($\theta_1'$, 3B)} & 0.6259 & 0.6623 & 0.6977 & 0.5813 & 0.5621 \\
\textbf{Time-R1 ($\theta_1$, 3B)} & \uline{0.6476} & \uline{0.6815} & \textbf{0.7555} & 0.5938 & 0.5599 \\
\bottomrule
\end{tabular}
}
\end{table}

\subsubsection{Stage 2: Future Event Time Prediction}

\begin{wrapfigure}{r}{0.5\textwidth} 
 \centering
 \includegraphics[width=\linewidth]{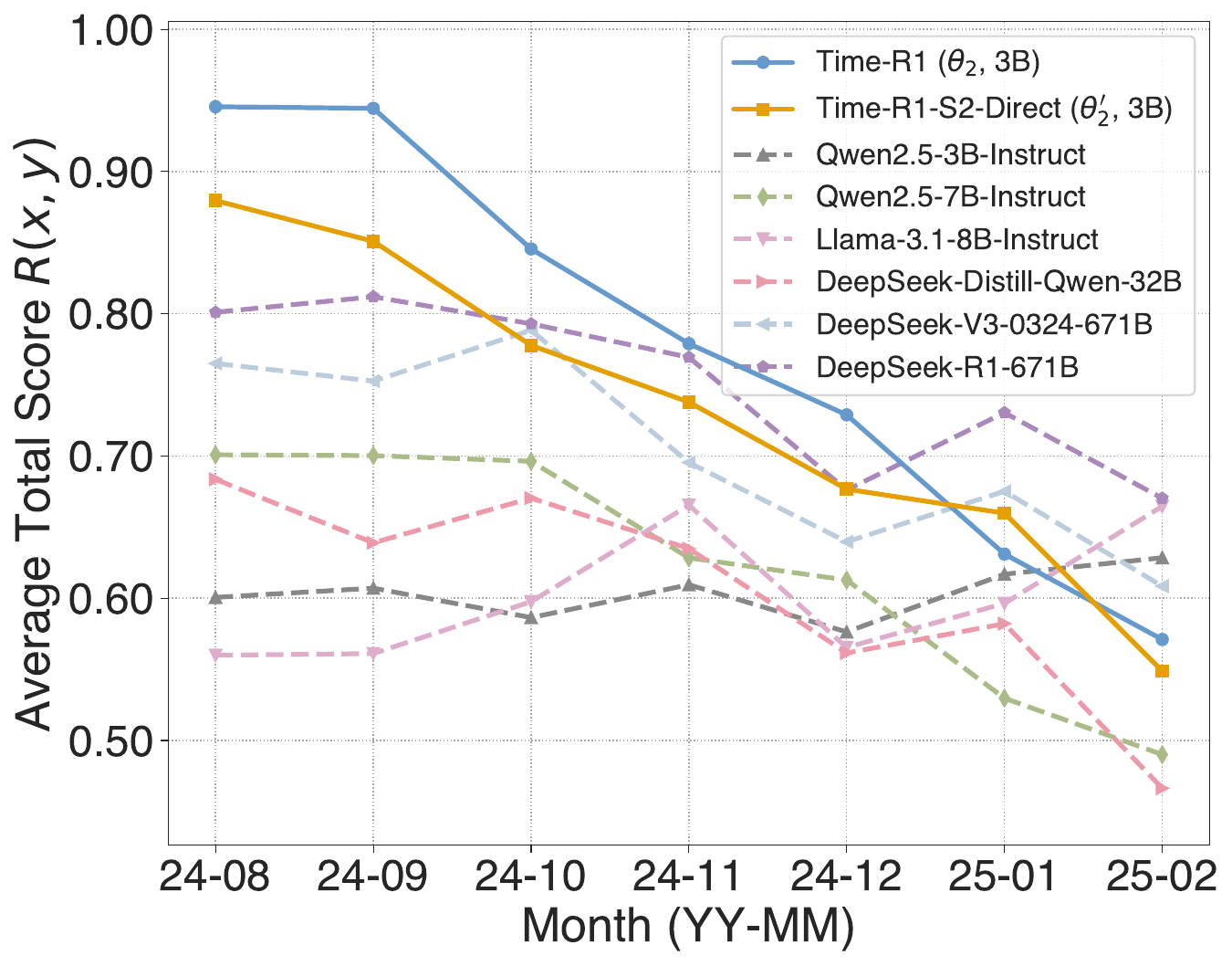} 
 \caption{Monthly Average Total Score $R(x, y)$ for Stage 2 Future Event Prediction (August 2024 - Feb 2025). Compares Time-R1 variants ($\theta_2$ and $\theta_2'$) against baselines. Evaluated with $\alpha=0.1$.}
 \label{fig:monthly_scores_wrapped}
 \vspace{-3mm} 
\end{wrapfigure}
Stage 2 equips models to predict event timing post-knowledge cutoff (2023). We assess our full pipeline and Stage 1's impact by evaluating two variants: \textbf{Time-R1 ($\theta_2$, 3B)} (full curriculum, \Cref{sec:experimental_design}) and an ablation model, \textbf{Time-R1-S2-Direct ($\theta_2'$, 3B)} (Stage 2 fine-tuning only, from base Qwen2.5-3B-Instruct, omitting Stage 1). Performance is compared against baselines for August 2024 - February 2025 predictions.

The overall Stage 2 performance, measured by Average Total Score $R(x,y)$ with strict evaluation ($\alpha=0.1$), is presented in \Cref{tab:stage2_overall_score_horizontal}. While models show clear improvement over Stage 1 Inference tasks, likely aided by a narrower prediction time span, further significant gains prove challenging. For instance, the DS-Qwen-32B model, despite its scale and specialized complex reasoning training, scores lower than some 3B models lacking such enhancements (\eg, the base Qwen2.5-3B-Instruct), underscoring the inherent difficulty of learning extrapolation and handling post-cutoff data.
Our primary model, Time-R1 ($\theta_2$, 3B), achieves the highest score. This strong performance, consistent across the prediction horizon (\Cref{fig:monthly_scores_wrapped}), shows it generally outperforming most baselines, including the much larger DeepSeek-R1-671B and DeepSeek-V3-671B models. This robust result strongly supports our hypothesis that specialized, staged temporal fine-tuning enables smaller models to achieve superior performance on challenging future prediction tasks. Furthermore, these findings highlight general LLM weaknesses in temporal reasoning and underscore the efficacy and necessity of our structured training framework. The foundational understanding from Stage 1, combined with Stage 2's predictive skill development, underpins this strong near-future temporal reasoning (see \Cref{Dc:otherllm} for challenges facing standard LLMs).
The ablation model, Time-R1-S2-Direct ($\theta_2'$, 3B), also demonstrates solid performance, outperforming several baselines and indicating Stage 2 RL fine-tuning's standalone effectiveness. See more discussion on \Cref{sec:ablation_staged_curriculum}.
\begin{table}[h!]
    \vspace{-3mm}
    \centering
    \caption{\textbf{Stage 2 Future Event Prediction Performance (Overall).} Average Total Score $R(x, y)$ evaluated with $\alpha=0.1$. Higher scores are better. Best score is \textbf{bold}, second best is \uline{underlined}. $\theta_2$ checkpoint of Time-R1 is used.}
    \label{tab:stage2_overall_score_horizontal}
    \resizebox{\textwidth}{!}{%
    \begin{tabular}{@{}l ccc ccc cc@{}}
        \toprule
        \textbf{Metric} & Qwen2.5 & Qwen2.5 & Llama-3.1 & DS-Qwen & DS-V3 & DS-R1 & \textbf{Time-R1} & \textbf{Time-R1} \\
        & -3B & -7B & -8B & -32B & -671B & -671B & \textbf{($\theta_2'$, 3B)} & \textbf{($\theta_2$, 3B)} \\
        \midrule
        \textbf{Avg. Total Score} $\uparrow$ & 0.6036 & 0.6226 & 0.6015 & 0.5997 & 0.7036 & \uline{0.7503} & 0.7331 & \textbf{0.7780} \\
        \bottomrule
    \end{tabular}
    } 
    \vspace{-3mm}
\end{table}

\subsubsection{Stage 3: Creative Scenario Generation Quality}

Finally, we evaluate model generalization to generating plausible future scenarios—a task without explicit fine-tuning. \Cref{tab:stage3_monthly_avgmaxsim_final_reordered_v2} presents AvgMaxSim scores, quantifying the semantic plausibility of generated news scenarios against real news events (August 2024 - February 2025). Results demonstrate Time-R1 ($\theta_2$, 3B)'s strong generalization capability. It achieves the highest overall AvgMaxSim score, surpassing all baseline models, including the very large DeepSeek-V3-0324-671B and DeepSeek-R1-671B. Monthly scores for Time-R1 ($\theta_2$, 3B) also reveal consistently strong performance. This Stage 3 success, achieved without direct training on generation, underscores the S1+S2 curriculum's effectiveness in building robust, transferable temporal reasoning. These capabilities are significant for addressing research gaps in challenging future prediction and generation tasks and demonstrate practical application value. Our ablation model, Time-R1-S2-Direct ($\theta_2'$, 3B), also performs commendably, outperforming some baselines (further discussion in \Cref{sec:ablation_staged_curriculum}).

\begin{table}[h!]
\vspace{-3mm}
    \centering
    \caption{\textbf{Stage 3 Creative Scenario Generation Plausibility (AvgMaxSim Scores (\%)).} Compares semantic similarity of generated scenarios to real news events (August 2024 - Feb 2025). Higher scores indicate better plausibility. Best overall average is \textbf{bold}, second best is \uline{underlined}.} 
    \label{tab:stage3_monthly_avgmaxsim_final_reordered_v2} 
    \resizebox{\textwidth}{!}{%
    \begin{tabular}{l c ccc ccc c} 
        \toprule
        \textbf{Model} & \textbf{Avg. (\%)} & \multicolumn{7}{c}{\textbf{Monthly AvgMaxSim Scores (\%)}} \\ 
        \cmidrule(lr){3-9} 
         & $\uparrow$ & 24-08 & 24-09 & 24-10 & 24-11 & 24-12 & 25-01 & 25-02 \\
        \midrule
        Qwen2.5-3B-Instruct       & 47.66 & 47.27 & 46.89 & 47.39 & 48.57 & 48.77 & 47.76 & 46.94 \\
        Qwen2.5-7B-Instruct       & 47.59 & 46.99 & 49.78 & 46.18 & 48.53 & 46.91 & 48.88 & 45.83 \\
        Llama-3.1-8B-Instruct     & 47.96 & 48.99 & 50.03 & 47.42 & 46.21 & 47.06 & 48.01 & 48.03 \\
        DeepSeek-Distill-Qwen-32B & 47.12 & 46.58 & 46.78 & 47.94 & 47.04 & 48.40 & 47.30 & 45.81 \\
        DeepSeek-V3-0324-671B     & \uline{48.81} & 50.73 & 51.77 & 48.60 & 48.46 & 47.52 & 47.71 & 46.85 \\ 
        DeepSeek-R1-671B          & 47.46 & 47.55 & 49.64 & 47.29 & 45.29 & 47.85 & 47.30 & 47.31 \\
        \midrule 
        \textbf{Time-R1-S2-Direct ($\theta_2'$, 3B)} & 47.93 & 47.89 & 47.11 & 47.95 & 48.29 & 46.05 & 50.69 & 47.52 \\
        \textbf{Time-R1 ($\theta_2$, 3B)}  & \textbf{48.90} & 47.75 & 48.29 & 49.81 & 48.77 & 49.03 & 50.81 & 47.83 \\
        \bottomrule
    \end{tabular}
    } 
    \vspace{-3mm}
\end{table}

\subsection{Ablation Studies}
\label{sec:ablation_study}




\subsubsection{Impact of Dynamic Reward Mechanism}
\label{sec:ablation_dynamic_reward}

Our methodology (\Cref{sec:dynamic_alpha}) employs a dynamic reward mechanism during Stage 1 fine-tuning. This curriculum learning approach, with its phased adjustment of reward strictness (from lenient $\alpha_{\text{start}}=0.07$ to strict $\alpha_{\text{target}}=0.1$), is designed to mitigate cold-start challenges and guide the model towards robust performance on complex temporal tasks. We hypothesized this would lead to superior learning compared to a static, strict reward function.

The empirical results presented in \Cref{fig:stage1_training_curves} validate this hypothesis. The advantage of the dynamic reward curriculum is evident both in the Overall Total Score across all subtasks (\Cref{fig:stage1_training_curves}a) and in the specific Masked Time Entity Completion subtask (\Cref{fig:stage1_training_curves}b). For the overall performance, the full Time-R1 model achieves a consistently higher and more stable score than the fixed-reward ablation model. This performance gap is even more pronounced in the Completion subtask, where the fixed-reward model's progress begins to slow and plateau around a score of 0.70. In contrast, the curriculum-trained model continues to improve, achieving a significantly higher and more stable final score of over 0.75. This suggests that the curriculum's initial leniency and gradual transition to stricter evaluation criteria enable more effective exploration and learning, preventing convergence to a sub-optimal policy and leading to a better mastery of the task.

\subsubsection{Impact of Staged Curriculum Learning}
\label{sec:ablation_staged_curriculum}

To quantify the impact of our staged curriculum, particularly the foundational comprehension from Stage 1, we compared our full model, Time-R1 ($\theta_2$, 3B) (S1+S2 training), against Time-R1-S2-Direct ($\theta_2'$, 3B) (S2 training only).

The results unequivocally highlight the benefits of the full curriculum. In Future Event Time Prediction (Stage 2, \Cref{tab:stage2_overall_score_horizontal}, \Cref{fig:monthly_scores_wrapped}), Time-R1 ($\theta_2$, 3B) (0.7780) significantly outperformed Time-R1-S2-Direct ($\theta_2'$, 3B) (0.7331). This advantage persisted in Stage 3 Creative Scenario Generation (\Cref{tab:stage3_monthly_avgmaxsim_final_reordered_v2}), with scores of 48.90\% and 47.93\% respectively. These consistent gains demonstrate that the temporal logic and event-time mapping skills instilled by Stage 1 are crucial for achieving superior predictive accuracy and generative plausibility, validating our progressive learning approach.

Notably, Time-R1-S2-Direct ($\theta_2'$, 3B) still demonstrated commendable performance, surpassing several baselines and even the larger DeepSeek-V3-671B in Stage 2. This underscores the inherent effectiveness of our Stage 2 RL fine-tuning for enhancing temporal reasoning. However, the superior performance of Time-R1 ($\theta_2$, 3B) across both tasks confirms that the initial foundational stage is key to unlocking the model's full potential, enabling a more comprehensive development of temporal intelligence from fundamental understanding to advanced prediction and generalization.

%% file: 5_discussion.tex
\section{Discussion}
\label{sec:discuss}
This section delves into a detailed analysis of our proposed methodology, focusing on the impact of our reasoning process on response length, and the challenges standard LLMs face in advanced temporal tasks. Our findings provide empirical evidence supporting the benefits of specialized training regimes for comprehensive temporal intelligence in LLMs. Additional discussion on implementation settings (\eg, KL loss coefficients), as well as more generated examples like those shown in \Cref{fig:time_r1_examples}, is available in \Cref{Appendix:KL_loss,Appendix:cases}.

\subsection{Reasoning Process Matters, Not Just Response Length}

\label{Dc:process}
\begin{wrapfigure}{r}{0.5\textwidth} 
 \centering
 \includegraphics[width=\linewidth]{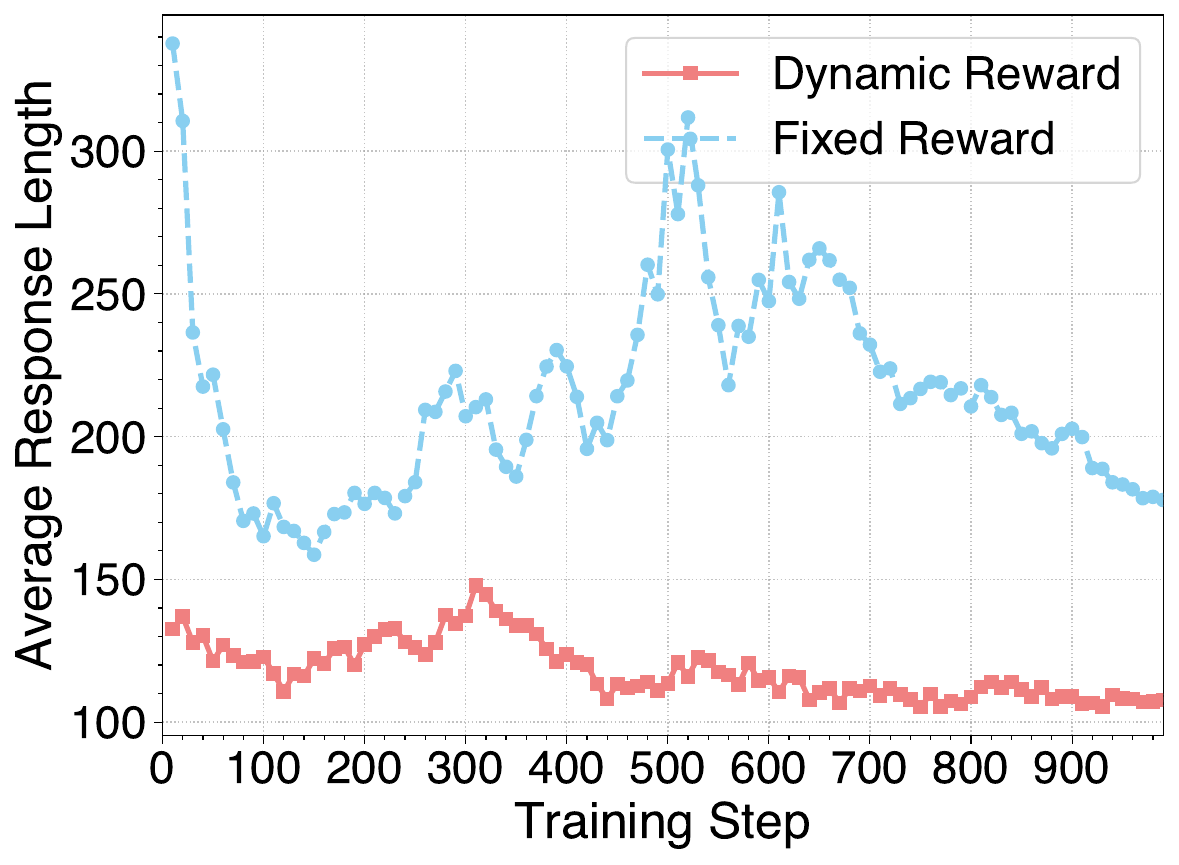} 
 \caption{\textbf{Impact of Dynamic Reward on Response Length.} The average response length (in tokens) across all Stage 1 tasks during training. The model trained with our full dynamic reward mechanism ("Dynamic Reward") produces consistently and significantly more concise outputs compared to the ablation model trained with a static, fixed reward ("Fixed Reward").}
 \label{fig:dynamic_reward_impact}
 \vspace{-3mm} 
\end{wrapfigure}

Developing effective LLMs requires not only accuracy but also efficient and concise responses. Unnecessarily long outputs can signify a less refined reasoning process and increase computational overhead. Our investigation reveals that our dynamic reward mechanism (\Cref{sec:dynamic_alpha}) achieves both higher accuracy and greater conciseness.

A combined analysis of our models' performance and output length provides compelling evidence for this. As established in \Cref{sec:ablation_dynamic_reward}, our dynamic reward curriculum leads to superior task performance (\Cref{fig:stage1_training_curves}). Simultaneously, \Cref{fig:dynamic_reward_impact} highlights the dramatic impact on average response length. The model trained with a fixed reward produces verbose outputs, averaging approximately 250 tokens. In stark contrast, the model trained with our dynamic reward mechanism generates significantly shorter responses, stabilizing at a much more efficient length of around 130 tokens.

This substantial reduction in length, achieved alongside superior task performance, strongly suggests that our curriculum fosters a more efficient and focused reasoning process. The model learns to achieve better outcomes without verbose outputs, implying a clearer, more direct approach to solving temporal tasks. Such conciseness is highly desirable, indicating a more refined understanding and leading to more interpretable and computationally efficient inferences.

\subsection{Challenges for Standard LLMs in Advanced Temporal Tasks}
\label{Dc:otherllm}

Standard Large Language Models (LLMs), including state-of-the-art reasoning-focused variants, exhibit commendable performance on foundational temporal tasks within their knowledge cutoff (Stage 1, \Cref{tab:stage1_temporal_reasoning_avg_total_score_decimal}). This is often attributable to their large scale and extensive pre-training, which can include significant mathematical and logical reasoning data. However, their capabilities are substantially challenged when faced with more advanced temporal tasks requiring extrapolation and nuanced future-oriented generalization.

Specifically, in Stage 2 Future Event Time Prediction (\Cref{tab:stage2_overall_score_horizontal}, \Cref{fig:monthly_scores_wrapped}) and Stage 3 Creative Scenario Generation (\Cref{tab:stage3_monthly_avgmaxsim_final_reordered_v2}), even powerful baselines like DeepSeek-R1-671B are outperformed by our significantly smaller Time-R1 ($\theta_2$, 3B). For instance, Time-R1 ($\theta_2$, 3B) achieved a leading score of 0.7780 in Stage 2 prediction (vs. DeepSeek-R1's 0.7503) and 48.90\% in Stage 3 generation (vs. DeepSeek-V3's 48.81\%). This disparity suggests that vast knowledge, large scale, or general reasoning prowess alone do not readily translate to proficiency in predicting future event timings or creatively generating plausible future scenarios. The relatively uniform and modest performance of baselines in Stage 3, in particular, highlights a general weakness in current LLM training methodologies to effectively generalize to future-oriented generation tasks.

In contrast, the success of our three-stage RL framework with Time-R1 ($\theta_2$, 3B) is notable. It not only excels in prediction but also demonstrates remarkable generalization to creative future scenario generation without any explicit fine-tuning on this generative task itself. This underscores the efficacy and robustness of our method in instilling a deeper, more transferable temporal understanding. These findings highlight the necessity for specialized training regimes like ours to cultivate comprehensive and practically useful temporal intelligence in LLMs.

%% file: 6_conclusion.tex
\section{Conclusion}
\label{sec:conclusion}

In this work, we introduced Time-R1, a 3B-parameter language model achieving comprehensive temporal reasoning—spanning understanding, prediction, and creative generation—through a novel, meticulously engineered three-stage reinforcement learning curriculum with a dynamic reward system. Strikingly, Time-R1 outperforms models over 200 times its size on challenging future event prediction and creative scenario generation tasks, exhibiting robust generalization to the latter even without task-specific fine-tuning. This success directly addresses a critical research gap concerning complex future-oriented tasks and demonstrates that our sophisticated, progressive RL approach enables smaller, efficient models to achieve superior temporal performance, offering a practical, scalable path towards truly time-aware AI with substantial application potential. To foster further research and development, we release our Time-Bench dataset and Time-R1 model checkpoints, envisioning future work on scalability and enhanced reasoning integration.

%% file: appendix.tex
\section{Experimental Configuration Details}
\label{app:config_details}

This appendix provides further details on the experimental setup and hyperparameter configurations used for the Reinforcement Learning (RL) fine-tuning of Time-R1, complementing the summary in Section~\ref{sec:exp_setup} of the main paper. Our experiments were conducted using the veRL framework \cite{sheng2024hybridflow}. 

\subsection{General Setup and Key Hyperparameters}
The base Large Language Model (LLM) for all our experiments is Qwen2.5-3B-Instruct. The RL fine-tuning was performed using 4 NVIDIA A6000 GPUs. Key hyperparameters for the Group Relative Policy Optimization (GRPO) algorithm and the overall training process are summarized in \Cref{tab:hyperparameters_appendix_revised}.

\begin{table}[htbp]
  \centering
  \caption{Key hyperparameters for RL fine-tuning Time-R1.}
  \label{tab:hyperparameters_appendix_revised}
  \begin{tabular}{lc}
    \toprule
    \textbf{Parameter}                                      & \textbf{Value}                 \\
    \midrule
    \textit{General \& Model}                                &                                \\
    Base Model Name                                         & Qwen2.5-3B-Instruct   \\
    Number of GPUs                                          & 4                     \\
    \midrule
    \textit{Data \& Batching}                                 &                                \\
    train batch size (Global)                        & 128                   \\
    GRPO mini batch size  & 64                    \\
    GRPO micro batch size          & 16                    \\
    max prompt length                                & 1024 tokens           \\
    max response length                              & 1024 tokens           \\
    \midrule
    \textit{Optimizer (Actor Model)}                          &                                \\
    learning rate                      & $2 \times 10^{-6}$    \\
    warmup style            & cosine                \\
    warmup steps   & 20                  \\
    \midrule
    \textit{GRPO Algorithm \& Rollout}                        &                                \\
    kl loss coef ($\beta$)        & 0.001                 \\
    rollout.n ($K$)                     & 5                     \\
    \bottomrule
  \end{tabular}
\end{table}

\subsection{Stage-Specific Training Configurations}
The multi-stage training of Time-R1 involved specific durations and checkpointing strategies for each stage, as outlined below. For both stages, checkpoints were selected based on the highest achieved score on the respective test set.

\textbf{Stage 1 (Comprehension):} This stage implemented our dynamic reward curriculum (detailed in \Cref{sec:dynamic_alpha}) and was divided into three phases:
\begin{itemize}
    \item \textbf{Phase 1} (Foundational Logic; Easy Timestamp Inference): Trained for 100 steps.
    \item \textbf{Phase 2} (Exploration; Full Task Suite, Mixed Difficulty): Trained for 500 steps.
    \item \textbf{Phase 3} (Transition to Strict Evaluation; Full Task Suite): Trained for 1000 steps.
\end{itemize}
Throughout Stage 1, evaluations on the test set were performed every 10 training steps, and model checkpoints were saved every 20 training steps. The best-performing checkpoint on the test set from each phase was used to initialize the subsequent phase or, for Phase 3, served as the final Stage 1 model ($\theta_1$).

\textbf{Stage 2 (Prediction):} This stage focused on future event time prediction:
\begin{itemize}
    \item Trained for 100 steps.
\end{itemize}
During Stage 2, both model checkpointing and test set evaluations occurred every 10 training steps. The checkpoint yielding the highest test score was selected as the final Stage 2 model ($\theta_2$).

These tailored configurations allowed for progressive and adaptive learning, ensuring that Time-R1 developed foundational understanding before advancing to more complex predictive tasks.

\section{Dataset Construction and Details}
\label{app:dataset_construction}

This appendix provides further details on the datasets used for training and evaluating Time-R1, supplementing the descriptions in Sections~\ref{sec:stage1} and \ref{sec:stage2}. 

\subsection{New York Times (NYT) Corpus Curation}
\label{app:nyt_corpus_details} 

The primary data source for our research is a corpus constructed from New York Times articles, utilizing publicly available information accessed via the NYT Archive API\footnote{\url{https://developer.nytimes.com/docs/archive-product/1/overview}}. For each article, we extracted key fields including the headline, abstract, publication date, and the ``news desk'' (thematic section).

We collected over 200,000 English-language NYT articles, with publication dates spanning from January 2016 to February 2025. To ensure the relevance of the articles to common temporal reasoning scenarios and current events, we selectively curated content from the following news desks: ``Politics'', ``National'', ``Washington'', ``U.S.'', ``Business'', ``SundayBusiness'', ``RealEstate'', ``Foreign'', ``World'', ``Metro'', ``Science'', ``Health'', ``Climate'', ``Opinion'', and ``OpEd''. Other news desks were excluded as they were found to reference current events less frequently.

This extensive NYT corpus was utilized for several distinct purposes within our framework:
\begin{itemize}
    \item \textbf{Stage 1 (Comprehension) Training Data:} Articles published from January 2016 to December 2023 were used to train the foundational temporal understanding capabilities of Time-R1 (see Section~\ref{sec:stage1} for Stage 1 details).
    \item \textbf{Stage 2 (Prediction) Real News Training Data:} A subset of articles from January 2024 to July 2024 served as real-world news data for the initial phase of Stage 2 training.
    \item \textbf{Stage 2 (Prediction) Real News Test Data:} Articles from August 2024 to February 2025 were held out and used as the real-news test set ($\mathcal{D}_{\text{test}}^{\smash{(2)}}$) for evaluating future event prediction performance.
\end{itemize}
In our task formulations, an event $E$ is typically represented by its headline $h$ and abstract $a$, i.e., $E=(h,a)$.

\subsection{Synthetic Data Generation for Future Event Prediction Training}
\label{app:synthetic_data_generation} 

To train Time-R1 for predicting events in future months (specifically, August 2024 to February 2025) without encountering data leakage from the real-news test period, we employed a data synthesis strategy as detailed in Section~\ref{sec:stage2}. This process utilized the DeepSeek-V3 model with a knowledge cutoff in July 2024.

The methodology for generating synthetic news articles was as follows:
\begin{itemize}
    \item \textbf{Targeted News Desk Distribution:} The generation aimed to reflect a historical distribution of articles across various news desks, based on NYT data prior to 2024. The primary target desk distribution used to guide generation proportions was:
    \begin{quote}
        Foreign: 20.8\%; Business: 16.5\%; OpEd: 14.2\%; National: 10.9\%; Washington: 9.6\%; Metro: 8.6\%; Politics: 5.5\%; Science: 4.6\%.
    \end{quote}
    \item \textbf{Few-Shot Prompting Strategy:} To generate content for a specific target future month (between August 2024 and February 2025) and a designated news desk, the DeepSeek-V3 model was prompted using a few-shot learning approach. Each prompt contained three real news headlines and abstracts from the \textit{same} news desk, randomly sampled from articles published between May 2024 and July 2024.
    \item \textbf{Generation Task:} For each such prompt, DeepSeek-V3 was instructed to generate six distinct synthetic news items (each comprising a headline and an abstract) relevant to the specified future month and news desk, learning from the style and content of the provided examples.
    \item \textbf{Output Distribution:} The selection and aggregation of these generated articles were managed so that the overall proportion of news items per desk for each future month in the synthetic training set ($\mathcal{D}_{\text{train}}^{\smash{(2)}}$) approximately mirrored the historical desk distribution detailed above.
\end{itemize}
This synthetic dataset provided the necessary training signals for the model to learn to predict events beyond its real-data cutoff while strictly ensuring no overlap with the real-news test data from the same period. The volume of this synthetic data for August 2024 - February 2025 was about half that of the real news data used for January 2024 - July 2024 in the Stage 2 training.

\section{Detailed Stage 1 Learning Curves and Analysis}
\label{app:stage1_curves}

This section provides a more detailed look at the learning dynamics during Stage 1 (Comprehension), complementing the summarized performance presented in \Cref{tab:stage1_temporal_reasoning_avg_total_score_decimal} of \Cref{sec:Stage1_results}. We present the training curves for all four fundamental temporal subtasks---Timestamp Inference, Time-Difference Estimation, Event Ordering, and Masked Time Entity Completion---specifically focusing on their progression throughout Phase 2 and Phase 3 of our dynamic reward curriculum (see \Cref{sec:dynamic_alpha} for details on the curriculum phases). Additionally, we illustrate the evolution of the inconsistency penalty factor ($P_{\text{incon}}$) for the Time-Difference Estimation and Event Ordering tasks during Phase 2, highlighting the model's improving adherence to logical and mathematical consistency.

\begin{figure}[htbp]
  \centering
  \includegraphics[width=\textwidth]{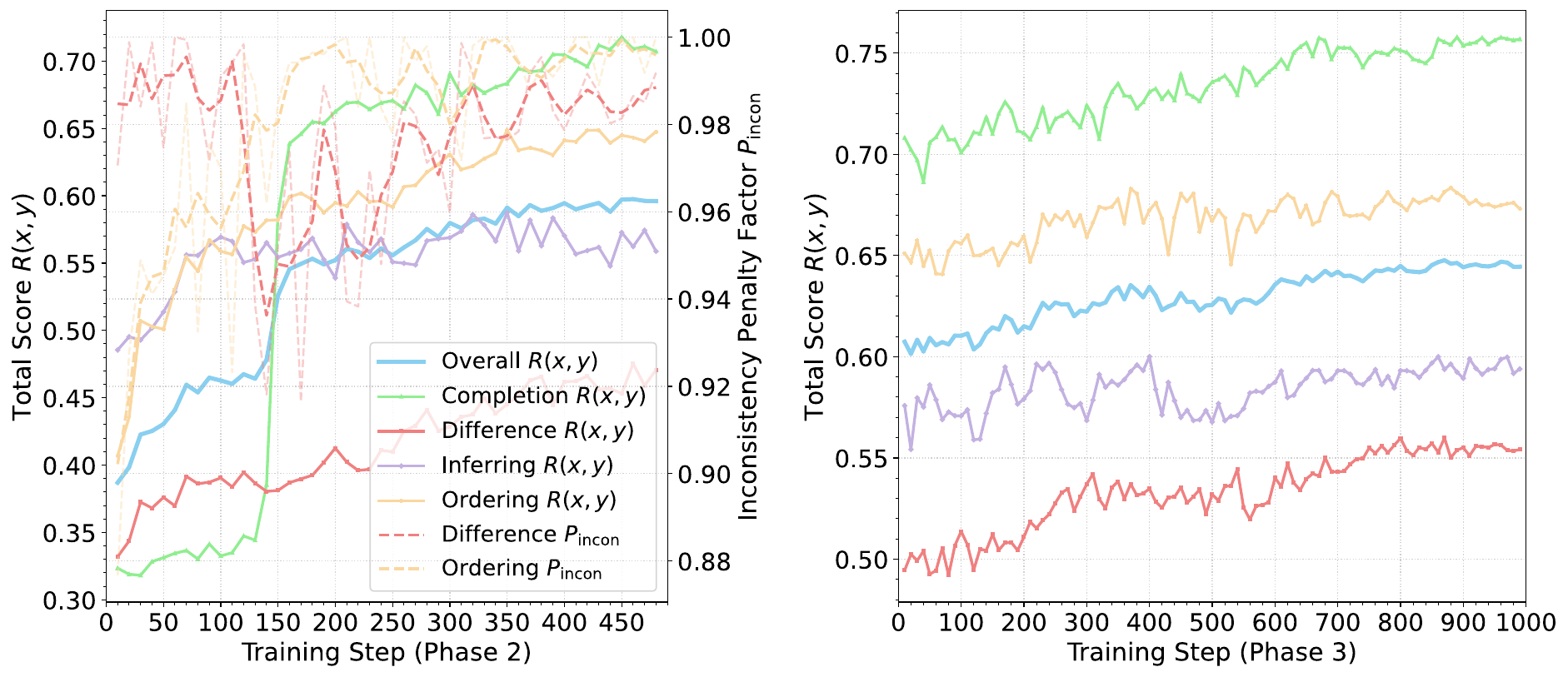} 
  \caption{Learning curves for Stage 1 subtasks during (Left) Phase 2 and (Right) Phase 3 of the dynamic reward curriculum. The left plot also shows the Inconsistency Penalty Factor ($P_{\text{incon}}$) for Time-Difference Estimation and Event Ordering tasks on the right y-axis during Phase 2.}
  \label{fig:stage1_learning_curves}
\end{figure}

The learning curves depicted in \Cref{fig:stage1_learning_curves} offer several key insights into the effectiveness of our methodology. 
Firstly, the steady increase and eventual convergence of the total scores ($R(x,y))$) across all subtasks in both Phase 2 and Phase 3 underscore the benefits of our dynamic reward design and curriculum learning strategy. This carefully structured approach enables the model to progressively master complex temporal logic, gradually adapting from more lenient to stricter evaluation criteria. As noted in \Cref{sec:Stage1_results}, this robust Stage 1 performance allows our 3B Time-R1 model to surpass numerous baseline models, many of which are ten to over two hundred times larger in parameter count (\Cref{tab:stage1_temporal_reasoning_avg_total_score_decimal}). Such strong foundational capabilities in temporal comprehension are crucial and deliberately engineered to provide a solid grounding for the subsequent, more demanding future-oriented tasks in Stage 2 (Prediction) and Stage 3 (Generation).

Secondly, the trends observed for the inconsistency penalty factors ($P_{\text{incon}}$) for the Time-Difference Estimation and Event Ordering tasks during Phase 2 (left plot of \Cref{fig:stage1_learning_curves}, dashed lines) are particularly revealing. The increasing values of $P_{\text{incon}}$ (approaching 1.0) indicate that the model is effectively learning to minimize inconsistencies in its responses. For instance, in Time-Difference Estimation, it learns to ensure that the explicitly stated time difference aligns with the difference calculated from its inferred dates for the two events. Similarly, for Event Ordering, the model becomes better at ensuring the stated order of events is consistent with the chronological sequence implied by its inferred dates for those events. This demonstrates that the penalty mechanisms detailed in \Cref{sec:task_specific_rewards} successfully guide the model not just towards task-specific accuracy but also towards generating responses that are logically coherent and mathematically sound, a critical aspect of true temporal understanding. By the commencement of Phase 3, these consistency factors are generally high, allowing the training to focus further on refining accuracy under strict evaluation.

Overall, these detailed learning dynamics from Stage 1 highlight the efficacy of our curriculum in building both accurate and logically consistent temporal reasoning, providing the essential groundwork for Time-R1's advanced capabilities in navigating future temporal challenges.

\section{Further Discussion on Implementation Settings} 
\label{Appendix:KL_loss} 

This appendix elaborates on specific implementation settings, focusing on the impact of the KL loss coefficient on model response length and the overall stability of our training framework with respect to various hyperparameter changes. These details supplement the primary configurations presented in \Cref{tab:hyperparameters_appendix_revised}.

\subsection{Impact of KL Loss Coefficient on Response Length}
The Group Relative Policy Optimization (GRPO) objective function, as defined in \Cref{eq:grpo_objective_final}, incorporates a KL divergence term $\mathbb{D}_{KL}[\pi_{\theta}(\cdot|x) || \pi_{ref}(\cdot|x)]$ scaled by a coefficient $\beta$. This term penalizes deviations of the current policy $\pi_{\theta}$ from a reference policy $\pi_{ref}$, encouraging smoother and more stable policy updates. The magnitude of $\beta$ directly influences the strength of this regularization.

\begin{figure}[htbp]
  \centering
  \includegraphics[width=0.7\textwidth]{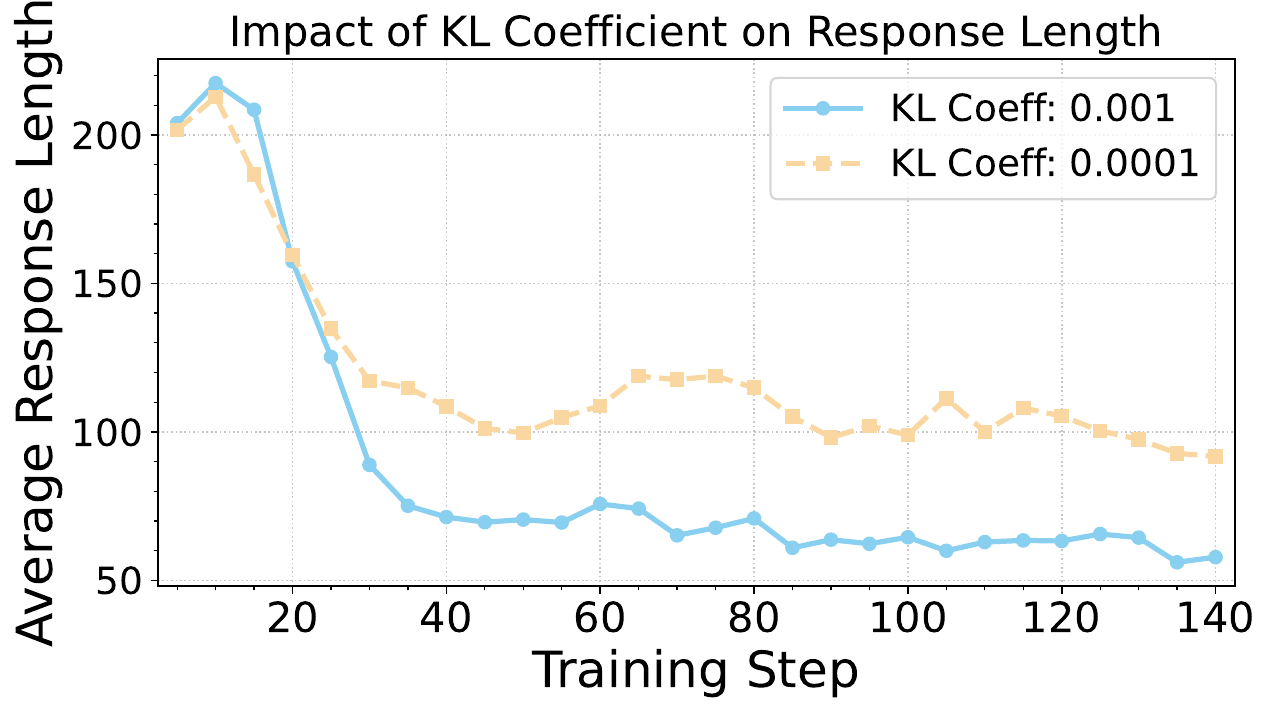} 
  \caption{Impact of different KL loss coefficients ($\beta$) on the average response length during training. A lower coefficient (0.0001) leads to longer average responses compared to the default setting (0.001).}
  \label{fig:kl_coeff_response_length}
\end{figure}

As illustrated in \Cref{fig:kl_coeff_response_length}, a lower KL coefficient (e.g., $\beta = 0.0001$ compared to our default $\beta = 0.001$) reduces the penalty for deviating from the reference policy. This allows the model greater freedom to explore diverse generation strategies during training. A noticeable consequence of this increased exploration with a lower $\beta$ is an increase in the average length of the generated responses. However, our experiments indicated that while the response lengths varied, the overall performance scores on the test sets remained largely comparable across these KL coefficient settings. This suggests that while the KL coefficient can influence stylistic aspects of the generation, such as verbosity, the core temporal reasoning capabilities learned by the model are robust within this range of $\beta$ values.

\subsection{Framework Stability under Hyperparameter Variations}
Beyond the KL coefficient, we investigated the sensitivity of Time-R1's performance to variations in other key hyperparameters relative to our main configuration detailed in \Cref{tab:hyperparameters_appendix_revised}. These variations included:
\begin{itemize}
    \item Increasing the number of rollout responses ($K$) from 5 to 8 and 11.
    \item Adjusting the sampling temperature from 1.0 to 1.2.
    \item Modifying the learning rate from $2 \times 10^{-6}$ to $5 \times 10^{-6}$ (increase) and $1 \times 10^{-6}$ (decrease).
    \item Increasing the GRPO micro batch size from 16 to 32.
    \item Varying the GRPO mini batch size from 64 to 128 (increase) and 32 (decrease).
\end{itemize}
Across these diverse hyperparameter modifications, we observed that the performance of Time-R1 on our test sets remained largely consistent, with no significant degradation in scores. This robustness to moderate changes in key training parameters underscores the overall stability and reliability of our proposed three-stage RL framework and GRPO optimization setup. Such stability is advantageous, suggesting that the framework is not overly sensitive to precise hyperparameter tuning, which can be beneficial for practical application and further development.

\section{Additional Generated Examples of Time-R1}
\label{Appendix:cases} 

This appendix presents additional generated examples from our Time-R1 model, supplementing Figure~\ref{fig:time_r1_examples} and further illustrating its capabilities across Stage 1 (Comprehension), Stage 2 (Prediction), and Stage 3 (Generation). These examples showcase the model's structured reasoning process (within \texttt{<think>...</think>} tags) and its final outputs (within \texttt{<answer>...</answer>} tags), alongside ground truth information and achieved scores. The detailed prompts used to elicit these responses are available in Appendix~\ref{app:prompt}. Our analysis highlights how Time-R1 demonstrates comprehensive temporal reasoning by effectively understanding context, making logical inferences, and generating plausible future-oriented content.


\subsection{Example: Stage 1 - Timestamp Inference}
\label{app:example_stage1_inference}

This example demonstrates Time-R1's ability to infer the publication date of a news article by reasoning about the real-world events mentioned.

\begin{table}[htbp]
  \centering
  \caption{Example of Timestamp Inference by Time-R1 $\theta_1$.}
  \label{tab:example_inference_revised_aligned} 
  \begin{tabularx}{\textwidth}{@{}lX@{}} 
    \toprule
    \textbf{Aspect} & \textbf{Details} \\
    \midrule
    Task & Stage 1: Timestamp Inference \\
    Input Headline & How to Build a Better Child Care System \\
    Input Abstract & The coronavirus has highlighted how necessary good child care is. \\
    \addlinespace 
    \begin{tabular}[t]{@{}l@{}}Model Reasoning \\ \texttt{<think>...</think>}\end{tabular} & 
    \texttt{The article discusses how the coronavirus has highlighted the necessity of a better child care system. Given that the \textbf{coronavirus outbreak began in January 2020} and has had a significant impact on child care systems around the world, it is reasonable to infer that the article is describing efforts to improve the child care system in response to the coronavirus pandemic. The article likely covers the events happening during the \textbf{spring of 2020}, when many countries were implementing policies to support child care during the pandemic.} \\
    \addlinespace
    \begin{tabular}[t]{@{}l@{}}Model Answer \\ \texttt{<answer>...</answer>}\end{tabular} & 
    \texttt{2020-04} \\
    \midrule
    Ground Truth (Event Pub Date) & 2020-05 \\
    Total Score & 1.005 \\ 
    \bottomrule
  \end{tabularx}
\end{table}

\paragraph{Analysis}
In this instance, Time-R1 correctly associates the discussion about child care systems with the coronavirus pandemic. It leverages its knowledge that the outbreak began in early 2020 and reasons that related articles discussing systemic responses would likely appear in the subsequent months. The inferred date \texttt{2020-04} is very close to the ground truth \texttt{2020-05}, showcasing accurate temporal localization based on contextual understanding of significant world events. The high score reflects this accuracy and proper formatting.

\subsection{Example: Stage 1 - Masked Time Entity Completion}
\label{app:example_stage1_completion_revised} 

This example illustrates the model's capability to not only infer an event's primary date but also to fill in a masked temporal entity within the text, requiring a deeper semantic understanding.

\begin{table}[htbp]
  \centering
  \caption{Example of Masked Time Entity Completion by Time-R1 $\theta_1$.}
  \label{tab:example_completion_revised} 
  \begin{tabularx}{\textwidth}{@{}lX@{}} 
    \toprule
    \textbf{Aspect} & \textbf{Details} \\
    \midrule
    Task & Stage 1: Masked Time Entity Completion \\
    Input Headline & How Russian Spies Hid Behind Bitcoin in Hacking Campaign \\
    Input Abstract & Operatives used cryptocurrency to obscure their identities and pay for the infrastructure needed to steal emails from Hillary Clinton’s \texttt{<YEAR>} presidential campaign. \\
    \addlinespace
    \begin{tabular}[t]{@{}l@{}}Model Reasoning \\ \texttt{<think>...</think>}\end{tabular} & \texttt{The article is about the hacking campaign that took place during the \textbf{2016 presidential election}. Given that the hacking campaign involved stealing emails from the Hillary Clinton presidential campaign, it is reasonable to infer that the \textbf{hacking campaign took place during the 2016 election year}. The hacking campaign would have needed time to happen in order to steal the emails, and the use of cryptocurrency would have also needed time to develop and be implemented.} \\
    \addlinespace
    \begin{tabular}[t]{@{}l@{}}Model Answer \\ \texttt{<answer>...</answer>}\end{tabular} & \texttt{Event: 2018-06. Missing entity: 2016.} \\
    \midrule
    Ground Truth & Event Pub Date: 2018-07, Masked Entity: 2016 \\
    Total Score & 1.052 \\
    \bottomrule
  \end{tabularx}
\end{table}

\paragraph{Analysis}
Time-R1 successfully identifies the masked year as \texttt{2016} by connecting the context of "Hillary Clinton's presidential campaign" to the correct election cycle. Simultaneously, it infers the main event's date (the article's publication discussing these past activities) as \texttt{2018-06}, which is very close to the ground truth \texttt{2018-07}. This demonstrates its ability to distinguish between the time of the events discussed within the text (the 2016 campaign) and the time of the news reporting itself, showcasing a nuanced understanding of temporal references and context.

\subsection{Example: Stage 2 - Future Event Time Prediction}
\label{app:example_stage2_prediction_revised} 

This example showcases Time-R1's ability to predict the timing of future events by extrapolating from patterns and general knowledge.

\begin{table}[htbp]
  \centering
  \caption{Example of Future Event Time Prediction by Time-R1 $\theta_2$.}
  \label{tab:example_prediction_revised} 
  \begin{tabularx}{\textwidth}{@{}lX@{}} 
    \toprule
    \textbf{Aspect} & \textbf{Details} \\
    \midrule
    Task & Stage 2: Future Event Time Prediction \\
    Input Headline & The Olympics, a Triumph of Ambition, Lift France From Its Gloom \\
    Input Abstract & All sectors of society came together to achieve the nation’s dream for the Games. Success has shown that unity is possible, if often elusive. \\
    \addlinespace
    \begin{tabular}[t]{@{}l@{}}Model Reasoning \\ \texttt{<think>...</think>}\end{tabular} & \texttt{The \textbf{Olympics are known to happen every four years}, and the \textbf{most recent Olympics were held in 2020}. Given that the Olympics typically take at least two years to prepare for and the COVID-19 pandemic would have taken at least two years to resolve, it is reasonable to infer that the Olympics would happen relatively quickly after the end of the 2020 Olympics, which were \textbf{delayed by one year due to the COVID-19 pandemic}.} \\
    \addlinespace
    \begin{tabular}[t]{@{}l@{}}Model Answer \\ \texttt{<answer>...</answer>}\end{tabular} & \texttt{2024-08} \\
    \midrule
    Ground Truth (Event Pub Date) & 2024-08 \\
    Total Score & 1.100 \\
    \bottomrule
  \end{tabularx}
\end{table}

\paragraph{Analysis}
In this challenging future prediction task, Time-R1 correctly predicts the \texttt{2024-08} date for the Paris Olympics. Its reasoning demonstrates an understanding of typical event cycles ("every four years"), knowledge of recent past events (2020 Olympics and their delay), and the ability to synthesize this information to make an accurate future projection. This highlights its capacity for temporal extrapolation, a key component of comprehensive temporal intelligence. The perfect score reflects this accurate prediction.

\subsection{Example: Stage 3 - Creative Future Scenario Generation}
\label{app:example_stage3_generation_revised} 

This example illustrates Time-R1's capability for creative scenario generation, where it generates a plausible future news item for a given future date (January 2025 in this case), without explicit fine-tuning on this generative task. The quality is assessed by semantic similarity to actual news from that period.

\begin{table}[htbp]
  \centering
  \caption{Example of Creative Future Scenario Generation by Time-R1 $\theta_2$ (Target: January 2025).}
  \label{tab:example_generation_revised} 
  \begin{tabularx}{\textwidth}{@{}lX@{}} 
    \toprule
    \textbf{Aspect} & \textbf{Details} \\
    \midrule
    Task & Stage 3: Creative Future Scenario Generation \\
    Target & News for January 2025 (Technology/Market theme implied by output) \\
    \addlinespace
    \multicolumn{2}{@{}l@{}}{\textbf{Time-R1 Generated News}} \\ 
    \textit{Headline:} & Global Tech Stocks Face Pressure as AI Innovations Begin to Shape the Market \\
    \textit{Abstract:} & The introduction of new artificial intelligence (AI) technologies is putting pressure on the tech stock market. Analysts predict that companies that can effectively integrate AI into their operations will see significant gains, while others may struggle to keep up. \\
    \addlinespace
    \multicolumn{2}{@{}l@{}}{\textbf{Comparable Real News}} \\
    \textit{Headline:} & China’s A.I. Advances Spook Big Tech Investors on Wall Street \\
    \textit{Abstract:} & Stiffer competition for the tech giants at the forefront of the artificial intelligence boom prompted investors to reassess the companies’ sky-high valuations. \\
    \midrule
    Semantic Similarity Score & 0.6731 \\
    \bottomrule
  \end{tabularx}
\end{table}

\paragraph{Analysis}
For the future date of January 2025, Time-R1 generated a plausible news scenario about AI's impact on tech stocks. This generated content is thematically coherent and discusses a relevant potential development in the technology and market sectors. When compared to a real news headline from a similar period that also discusses AI and tech investors, it achieves a notable semantic similarity score (0.6731). This demonstrates Time-R1's ability to not just predict dates, but to creatively generate contextually relevant and plausible future narratives, showcasing a strong generalization of its learned temporal understanding and reasoning skills. This ability to generate novel, coherent future content is a hallmark of advanced temporal intelligence.

\section{Illustration of Length and Repetition Penalty Efficacy}
\label{app:repete} 

In Section~\ref{sec:universal_bonuses_penalties} (Common Bonuses and Penalties), we introduced the Length and Repetition Penalty ($P_{\text{len\_rep}}$), designed to discourage overly verbose or repetitive model outputs. We noted that this mechanism has proven particularly effective. This section provides an illustrative example of the type of repetitive reasoning that the $P_{\text{repetition}}$ component of this penalty targets, thereby guiding the model towards more efficient and varied responses.

\begin{table}[htbp]
  \centering
  \caption{Example Illustrating Repetitive Reasoning Targeted by the $P_{\text{repetition}}$ Penalty.}
  \label{tab:example_repetition}
  \begin{tabularx}{\textwidth}{@{}lX@{}} 
    \toprule
    \textbf{Aspect} & \textbf{Details} \\
    \midrule
    Task & Time Prediction (Stage 2 Example) \\
    Input Headline & After Thursday’s Debate, Conservative Media Finds It Hard Not to Gloat \\
    Input Abstract & Many conservative commentators have said President Biden’s performance during the debate was a sweeping validation of the alarm bells they’d been ringing for years about his age. \\
    \addlinespace
    \begin{tabular}[t]{@{}l@{}}Model Reasoning \\ \texttt{<think>...</think>}\end{tabular} & \texttt{The article is about the debate between President Biden and other candidates in the Republican primary election. \textbf{Given that elections usually take several weeks to several months to be resolved, it is reasonable to infer that the article is about describing preparations for the debate.} \textbf{Given that elections usually take several weeks to several months to be resolved, it is likely that the article is about describing preparations for the debate.}} \\ 
    \addlinespace
    \begin{tabular}[t]{@{}l@{}}Model Answer \\ \texttt{<answer>...</answer>}\end{tabular} & \texttt{2024-06} \\
    \midrule
    Ground Truth (Event Pub Date) & 2024-07 \\
    \bottomrule
  \end{tabularx}
\end{table}

\paragraph{Analysis and Impact of Penalties}
The model's reasoning process shown in \Cref{tab:example_repetition} exhibits a clear instance of repetition: the core phrase ``Given that elections usually take several weeks to several months to be resolved, it is reasonable/likely to infer that the article is about describing preparations for the debate'' appears twice with only minor variation. This form of redundancy would be directly addressed by the $P_{\text{phrase\_repeat}}$ component within our $P_{\text{repetition}}$ penalty (as defined in \Cref{sec:universal_bonuses_penalties}). 

By applying such penalties, the $P_{\text{len\_rep}}$ mechanism actively discourages the model from generating verbose or repetitive content. This not only improves the conciseness of the output but also pushes the model to explore more diverse and efficient reasoning pathways. The consistent application of these universal bonuses and penalties, including those for length and various forms of repetition (word, phrase, n-gram diversity), is therefore instrumental in achieving the well-formed, succinct, and accurate responses demonstrated in the examples throughout \Cref{Appendix:cases}. It ensures that Time-R1's advanced temporal reasoning is communicated clearly and effectively, without being undermined by a tendency towards unnecessary verbosity or redundancy.

\section{Limitations}
\label{sec:limitations}


Firstly, while our work introduces Time-Bench, a large-scale, open-source dataset designed to facilitate comprehensive temporal reasoning research, a potential limitation lies in the scope of evaluation. Spanning a decade of news data and comprising over 200,000 examples across multiple temporal tasks, Time-Bench provides a robust benchmark for evaluating the capabilities demonstrated by Time-R1. However, validating the effectiveness and generalization capabilities of our model on a wider array of external temporal reasoning benchmarks and diverse datasets would further strengthen our findings and provide stronger evidence for the robustness of our proposed training framework.

Secondly, while our results demonstrate that smaller models can achieve strong performance on temporal tasks with specialized RL training, evidence from baseline comparisons also suggests that larger models generally exhibit higher capabilities. Due to resource constraints, we focused on demonstrating the efficacy of our approach on a 3B model to highlight cost-effective and rapid iteration potential. However, applying our three-stage RL framework to larger foundation models could likely yield even more significant performance gains, leveraging their inherently greater knowledge capacity. Our work primarily showcases the potential of the RL methodology, which we believe would scale positively with model size.

\section{Ethical Statement}
\label{sec:ethical_statement}

The development of Time-R1 and the Time-Bench dataset aims to advance research in temporal reasoning for AI. The dataset constructed from New York Times articles uses publicly available information through Archive api. While endowing models with future prediction and scenario generation capabilities has many beneficial applications, such as in planning and risk assessment, we acknowledge the potential for misuse, such as generating misleading future-oriented content. To address this, we believe that fostering an environment of transparency and critical use is paramount; users should be aware when content is AI-generated, particularly for probabilistic future scenarios, allowing for informed interpretation rather than uncritical acceptance. This approach, emphasizing clear attribution and critical engagement, combined with ongoing research into robust safeguards, is crucial for responsibly harnessing such powerful capabilities. Our model development did not involve human-derived private data beyond publicly archived news. The research was conducted with the intention of fostering a better understanding of AI's temporal intelligence, and we encourage responsible use and further investigation into safeguards for generative temporal models. The datasets and models will be released to the research community to promote transparency and further beneficial advancements in this domain.

\section{Prompts}
\label{app:prompt}

This appendix provides the detailed structure and content of the prompts used to guide our Large Language Model for each of the six temporal reasoning tasks evaluated in this work. Consistent with the methodology described in \Cref{sec:problem} (Structured Generation Process), all prompts employ a specific template designed to elicit chain-of-thought reasoning. This template includes system instructions directing the model to first articulate its reasoning process within `<think>...</think>` tags, followed by the final answer encapsulated in `<answer>...</answer>` tags. This structured approach aims to enhance the robustness of the model's reasoning and the interpretability of its outputs. The specific prompts for each task are detailed in the following subsections.

\subsection{Prompt for Timestamp Inference}
\label{app:prompt_timestamp_inference}

\begin{figure}[htbp] 
  \centering
  \includegraphics[width=\textwidth]{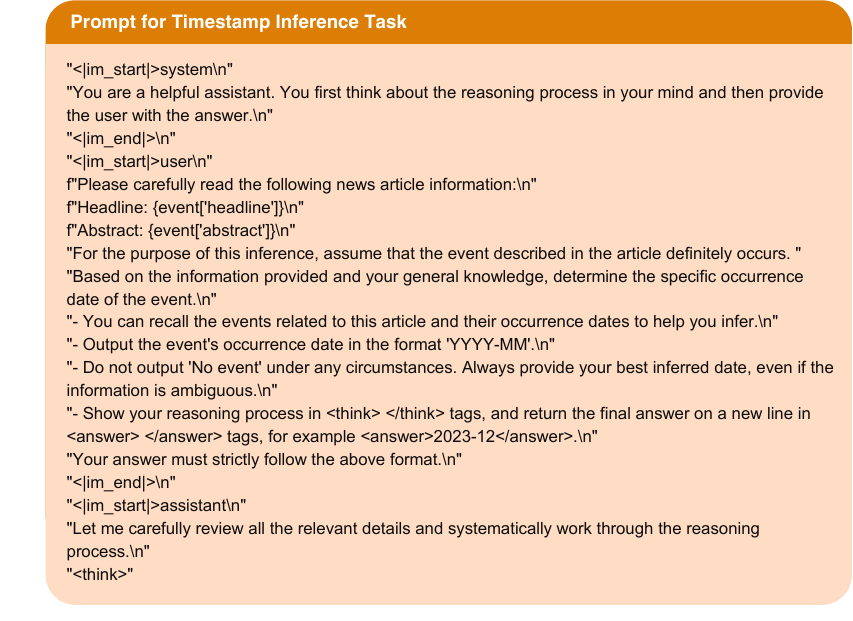} 
  \caption{Prompt for the Timestamp Inference task.}
  \label{fig:timestamp_inference_prompt_image}
\end{figure}

The Timestamp Inference task is one of the four fundamental temporal tasks in Stage 1 (Comprehension). It requires the model to infer the specific month and year (formatted as YYYY-MM) of an event based on its provided news headline and abstract. The detailed prompt given to the model for this task, including system messages, user input structure with placeholders for event details, and specific output formatting requirements, is shown in Figure~\ref{fig:timestamp_inference_prompt_image}.

\subsection{Prompt for Time-Difference Estimation}
\label{app:prompt_time_difference}

\begin{figure}[htbp]
  \centering
  \includegraphics[width=\textwidth]{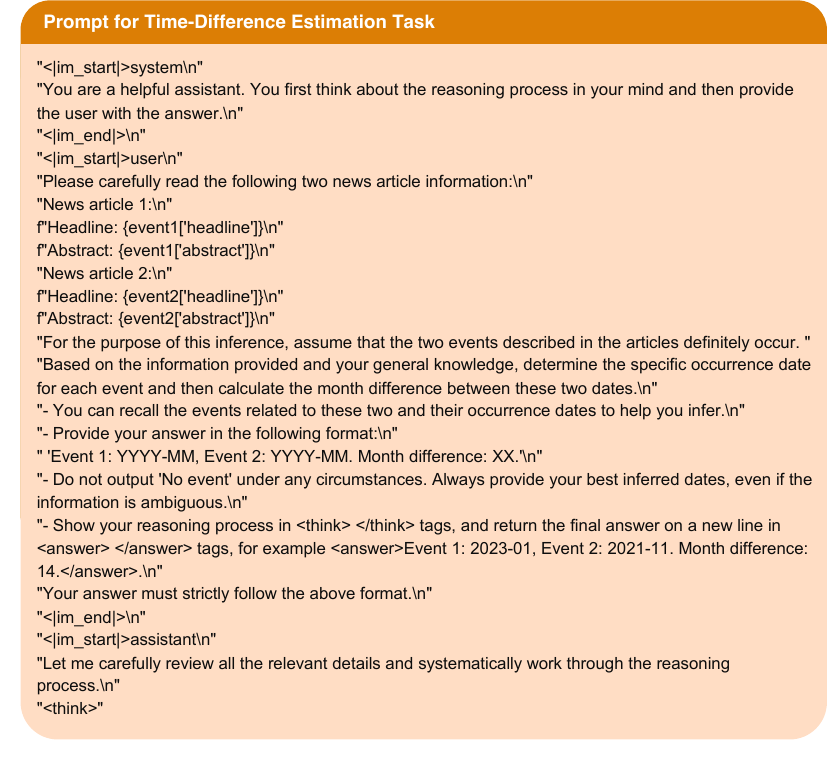} 
  \caption{Prompt for the Time-Difference Estimation task.}
  \label{fig:time_difference_prompt_image}
\end{figure}

The Time-Difference Estimation task is part of Stage 1 (Comprehension). It requires the model to first infer the specific dates of two separate events ($E_1$ and $E_2$) described by their news headlines and abstracts, and then to estimate the temporal gap (\ie, in months) between these two events. The detailed prompt guiding the model through this multi-step reasoning process is shown in Figure~\ref{fig:time_difference_prompt_image}.

\subsection{Prompt for Event Ordering}
\label{app:prompt_event_ordering}

\begin{figure}[htbp]
  \centering
  \includegraphics[width=\textwidth]{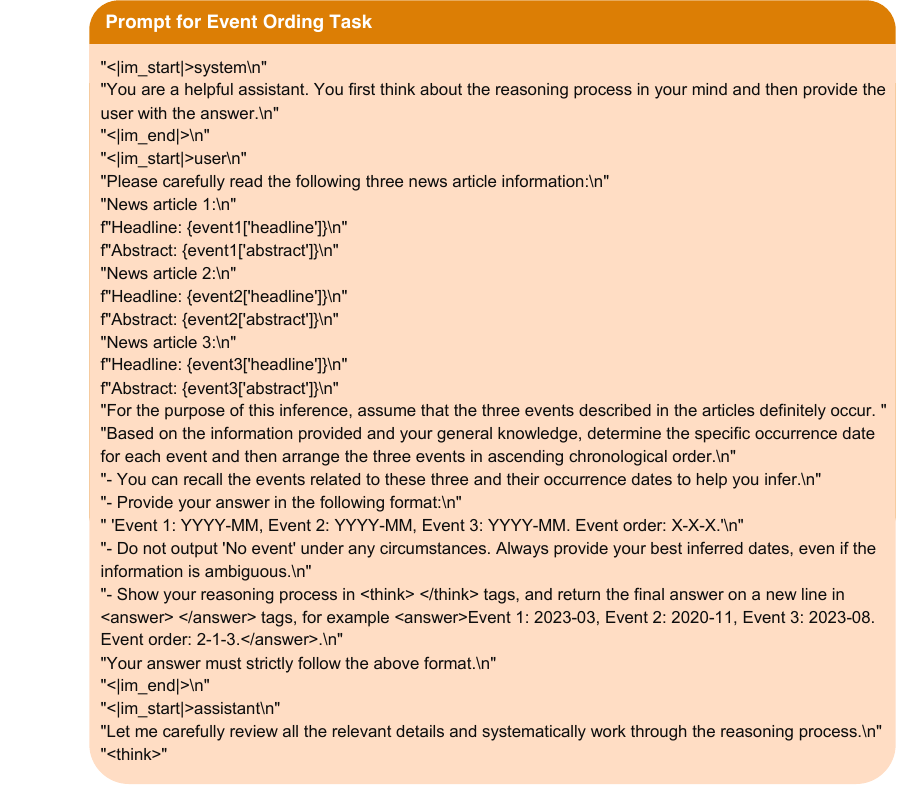} 
  \caption{Prompt for the Event Ordering task.}
  \label{fig:event_ordering_prompt_image}
\end{figure}

The Event Ordering task, also a component of Stage 1 (Comprehension), challenges the model to determine the correct chronological sequence of three distinct events ($E_1, E_2, E_3$) presented out of order. Similar to other Stage 1 tasks, the model is prompted to first infer the date of each event before determining their order. The prompt structure for this task is presented in Figure~\ref{fig:event_ordering_prompt_image}.

\subsection{Prompt for Masked Time Entity Completion}
\label{app:prompt_masked_time_entity}

\begin{figure}[htbp]
  \centering
  \includegraphics[width=\textwidth]{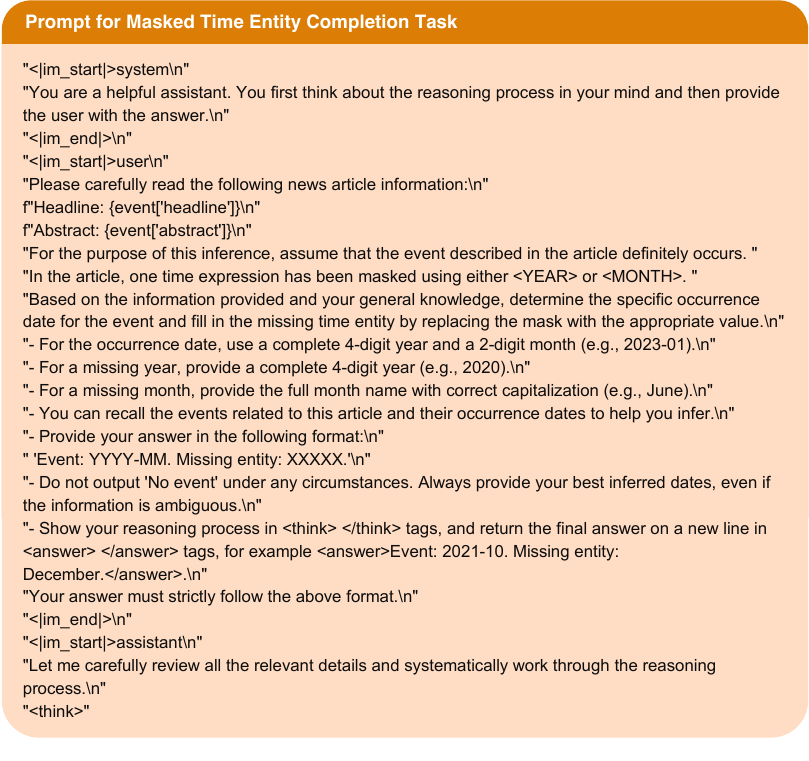} 
  \caption{Prompt for the Masked Time Entity Completion task.}
  \label{fig:masked_time_entity_prompt_image}
\end{figure}

The Masked Time Entity Completion task is the fourth fundamental task in Stage 1 (Comprehension). In this task, the model is given an event description ($E'$) containing a masked temporal expression (such as `<Year>` or `<Month>`) and is required to fill in the correct missing time entity, after first inferring the event's overall date. The specific prompt used to guide this completion process is shown in Figure~\ref{fig:masked_time_entity_prompt_image}.

\subsection{Prompt for Future Event Time Prediction}
\label{app:prompt_future_event_prediction}

\begin{figure}[htbp]
  \centering
  \includegraphics[width=\textwidth]{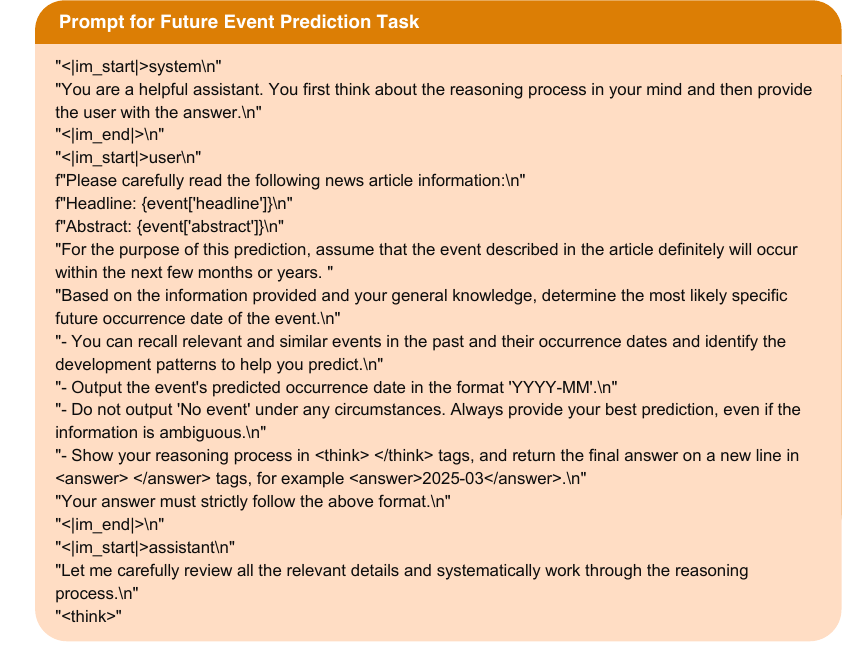} 
  \caption{Prompt for the Future Event Time Prediction task.}
  \label{fig:future_event_prediction_prompt_image}
\end{figure}

The Future Event Time Prediction task constitutes Stage 2 (Prediction) of our framework. Here, the model is tasked with predicting the specific future date (YYYY-MM) of a news event based on its extracted headline and abstract, focusing on events occurring after the model's initial knowledge cutoff. The prompt designed to elicit these future predictions is displayed in Figure~\ref{fig:future_event_prediction_prompt_image}.

\subsection{Prompt for Creative Future Scenario Generation}
\label{app:prompt_creative_scenario_generation}

\begin{figure}[htbp]
  \centering
  \includegraphics[width=\textwidth]{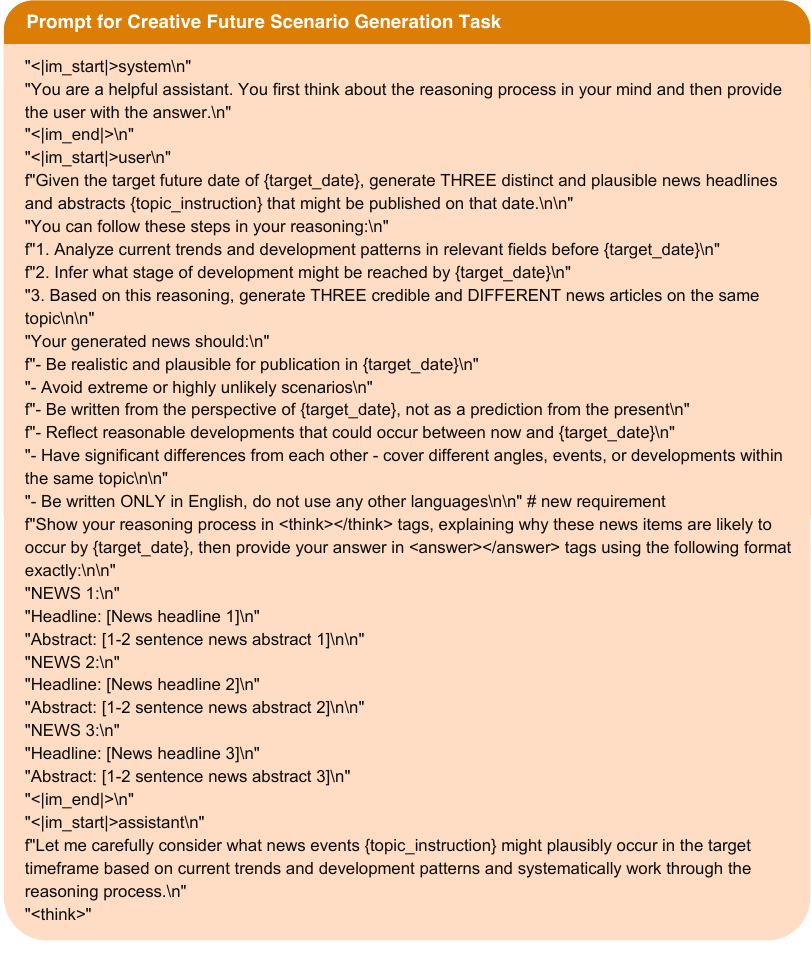} 
  \caption{Prompt for the Creative Future Scenario Generation task.}
  \label{fig:creative_scenario_generation_prompt_image}
\end{figure}

The Creative Future Scenario Generation task is the focus of Stage 3 (Generation). In this stage, the model leverages capabilities developed previously to generate plausible, hypothetical news event descriptions or headlines for a specified future date and thematic category (\eg, Business, Technology). This task evaluates the model's ability to creatively imagine coherent future events. The prompt used to guide this generative process is presented in Figure~\ref{fig:creative_scenario_generation_prompt_image}.

